\documentclass[10pt,twocolumn,letterpaper]{article}

\usepackage{wacv}              % To produce the CAMERA-READY version

\usepackage{graphicx}
\usepackage{amsmath}
\usepackage{amssymb}
\usepackage{booktabs}

\usepackage{tikz}
\usetikzlibrary{spy}

\usepackage[absolute]{textpos}
\usepackage[accsupp]{axessibility} % Improves PDF readability for those with disabilities.
\usepackage[pagebackref,breaklinks,colorlinks]{hyperref}

\usepackage[capitalize]{cleveref}
\crefname{section}{Sec.}{Secs.}
\Crefname{section}{Section}{Sections}
\Crefname{table}{Table}{Tables}
\crefname{table}{Tab.}{Tabs.}

\def\wacvPaperID{935} % *** Enter the WACV Paper ID here
\def\confName{WACV}
\def\confYear{2025}

\newcommand{\TODO}[1]{\textbf{\color{red}[TODO: #1]}}
\renewcommand{\TODO}[1]{}

\usepackage{amsmath}
\usepackage{amssymb}
\usepackage{cleveref}
\usepackage{subcaption}
\usepackage{booktabs}

\usepackage[normalem]{ulem}

\newcommand{\abs}[1]{\left\vert#1\right\vert}
\newcommand{\norm}[1]{\left\Vert#1\right\Vert}

\let\originalleft\left
\let\originalright\right
\renewcommand{\left}{\mathopen{}\mathclose\bgroup\originalleft}
\renewcommand{\right}{\aftergroup\egroup\originalright}

\definecolor{best}{rgb}{1, 0.7, 0.7}
\definecolor{second}{rgb}{1, 0.85, 0.7}
\definecolor{tab1}{rgb}{0.31,0.47,0.65}
\definecolor{tab2}{rgb}{0.95,0.56,0.17}

\newcommand{\imagewithzoom}[7]{\begin{tikzpicture}[spy using outlines={lens={scale=#7}, size=#6}]

\node[inner sep=0pt] at (0, 0)  {\includegraphics[width=\textwidth, keepaspectratio]{#1}};
        
\spy [red] on (#2,#3) in node at (#4,#5);
\end{tikzpicture}}

\usepackage{pifont}
\newcommand{\cmark}{\text{\ding{51}}}
\newcommand{\xmark}{\text{\ding{55}}}

\begin{document}

\title{
NeRFs are Mirror Detectors: Using Structural Similarity\\for Multi-View Mirror Scene Reconstruction with 3D Surface Primitives}

\author{
Leif Van Holland$^1$\quad Michael Weinmann$^2$\quad Jan U. Müller$^1$\quad Patrick Stotko$^1$\quad Reinhard Klein$^1$\\
$^1$University of Bonn\qquad $^2$Delft University of Technology\\
{\tt\small holland@cs.uni-bonn.de\qquad m.weinmann@tudelft.nl\qquad \{muellerj,stotko,rk\}@cs.uni-bonn.de}
}

\maketitle

%%%%%%%%% ABSTRACT
\begin{abstract}
    While neural radiance fields (NeRF) led to a breakthrough in photorealistic novel view synthesis, handling mirroring surfaces still denotes a particular challenge as they introduce severe inconsistencies in the scene representation.
    Previous attempts either focus on reconstructing single reflective objects or rely on strong supervision guidance in terms of additional user-provided annotations of visible image regions of the mirrors, thereby limiting the practical usability.
    In contrast, in this paper, we present NeRF-MD, a method which shows that NeRFs can be considered as mirror detectors and which is capable of reconstructing neural radiance fields of scenes containing mirroring surfaces without the need for prior annotations.
    To this end, we first compute an initial estimate of the scene geometry by training a standard NeRF using a depth reprojection loss.
    Our key insight lies in the fact that parts of the scene corresponding to a mirroring surface will still exhibit a significant photometric inconsistency, whereas the remaining parts are already reconstructed in a plausible manner.
    This allows us to detect mirror surfaces by fitting geometric primitives to such inconsistent regions in this initial stage of the training.
    Using this information, we then jointly optimize the radiance field and mirror geometry in a second training stage to refine their quality.
    We demonstrate the capability of our method to allow the faithful detection of mirrors in the scene as well as the reconstruction of a single consistent scene representation, and demonstrate its potential in comparison to baseline and mirror-aware approaches.
\end{abstract}

\vspace{-12pt}

\begin{textblock}{15}(3,25.8)
{\footnotesize\noindent\color{gray} © 2025 IEEE.  Personal use of this material is permitted.  Permission from IEEE must be obtained for all other uses, in any current or future media, including reprinting/republishing this material for advertising or promotional purposes, creating new collective works, for resale or redistribution to servers or lists, or reuse of any copyrighted component of this work in other works.
}
\end{textblock}

%%%%%%%%% BODY TEXT
\section{Introduction}
\label{sec:intro}

Being among the classical topics of computer vision and an essential prerequisite for numerous applications in autonomous systems, robotics, virtual prototyping, industrial inspection, advertisement, entertainment, architecture, cultural heritage, education and medical contexts, 3D reconstruction and modeling of scenes has gained a lot of attention for decades.
Despite the impressive progress that has been reached over the years, mirroring surfaces, such as highly reflective or glass surfaces that are common in architectural designs and interior spaces, present a significant challenge since they do not exhibit a multi-view consistent appearance themselves but rather reflect and potentially distort their environment.
Not adequately taking the detection and reconstruction of mirroring surfaces within respective scenes into account does not only affect the degree of realism perceived when inspecting reconstructed scenes but may also lead to severe implications for tasks like autonomous navigation, object tracking, quality control due to distractions induced by the mirroring surfaces.
Whereas the rapid progress in machine learning has led to the development of powerful techniques for neural scene representation and rendering~\cite{tewari2020state,tewari2022advances}, state-of-the-art approaches such as neural radiance fields (NeRF)~\cite{mildenhall2020nerf} and most of their extensions do not model physical reflection in their rendering pipeline but instead incorrectly interpret the reflection in the mirror as an independent virtual scene, providing the illusion of viewing the respective content through a window and thereby leading to an inaccurate reconstruction of the mirror's geometry with an incorrect multi-view consistent appearance.
Further methods~\cite{bi2020neural,boss2021nerd,srinivasan2021nerv,boss2022samurai,boss2021neural,jin2023tensoir,kuang2022neroic,verbin2022refnerf,zhang2022modeling,munkberg2022extracting,zhang2021nerfactor,zhao2022factorized} focused on the inference of a better scene understanding in terms of the decomposition of the observed appearance into object materials and illumination, thereby also considering reflection effects on surfaces. However, these approaches commonly assume that surfaces exhibit a certain degree of diffuse reflection for the initial recovery of the object surface, followed by the subsequent modeling of the specular component.
As a result, these methods encounter challenges when dealing with mirrors exhibiting pure specular reflection, primarily stemming from inaccuracies in estimating the mirrors' surface.
Handling mirroring surfaces within neural scene representations has been addressed based on decomposing the scene into different individually consistent neural fields such as the modeling of transmitted and reflected light components with separate neural fields in combination with geometric priors~\cite{guo2022nerfren,ma2023specnerf,verbin2024nerf} and scene modeling based on multiple parallel neural feature field~\cite{yin2023msnerf,wu2024neural,meng2024mirror,liu2024mirrorgaussian}, where in both approaches the resulting neural fields are blended for view synthesis.
Further NeRF extensions for specular surfaces~\mbox{\cite{zeng2023mirror}} rely on a single radiance field and standard path tracing~\cite{zhang2023nemf,mai2023neural} to trace rays into the ideal reflection direction after hitting a mirror, where respective normal directions and reflection probabilities for reflecting the rays are learned in the volumetric neural field. However, this comes at significantly increased computational costs regarding training times and rendering speed, thereby limiting the practicality for other sophisticated NeRF approaches.
Furthermore, the use of a decomposed scene representation in terms of components for diffuse and specular parts, separately modeled via NeRFs~\cite{yin2023msnerf} or 3D Gaussian splatting~\cite{meng2024mirror,liu2024mirrorgaussian}, have shown great promise.
However, these approaches rely on given segmentations of the mirroring surface parts in the input images and are restricted to planar mirroring objects.
TraM-NeRF~\cite{holland2023tramnerf} leveraged a transmittance-aware formulation of the rendering equation to explicitly model reflected radiance at mirror-like surfaces.
Furthermore, the authors also increase the efficiency in comparison to Monte-Carlo estimation approaches by using efficient importance sampling and transmittance computation.
Despite this progress, all the above-mentioned methods for handling mirroring surfaces rely on user annotations of masked regions that contain the mirroring objects, which is impractical to get for multi-view scenarios where a camera is used in lots of different views.
In this paper, we present a novel approach for neural scene representation and rendering for scenes containing mirroring surfaces that allows both the identification of mirroring surfaces and their reconstruction, thereby overcoming the respective limitations of previous approaches.
For this purpose, we start by inferring an initial geometry estimate for the scene based on the training of a standard radiance field~\cite{mildenhall2020nerf}, using a depth reprojection loss to improve the scene geometry~\cite{truong2023sparf}, and the subsequent extraction of surface points in the radiance field by rendering depths masks.
A key insight used in our approach is that the initial reconstruction faithfully reconstructs the scene, except for highly reflective surfaces.
In these regions, the rendered reconstruction will exhibit noticeable artifacts.
We use this fact to filter the surface points using a novel scoring scheme that combines local structural similarity compared to ground truth training data with the variance of the estimated depth.
We remove points with low scores from the point cloud and segment the remaining points into a set of primitive shapes, which in turn are used to explicitly model ray reflections \cite{holland2023tramnerf}, extending previous work to support differentiation with respect to the shapes' parameters.
This way, we can jointly optimize the radiance field and the initial shape estimates.

In summary, the key contributions of our work are:
\begin{itemize}
  \item We present a novel NeRF variant that allows the reconstruction of scenes containing mirroring surfaces without relying on priors in terms of annotations of regions covered by mirroring surfaces in the input images.
  \item We present a novel training strategy designed for scenes containing mirroring surfaces with an automated localization and reconstruction of mirroring surfaces. This is achieved by an initial NeRF training, where the result is leveraged to segment reflective surfaces as primitive shapes and use these in a joint NeRF model, optimizing both the volumetric representation and refine the primitive parameters during training.
  \item We demonstrate the potential of our approach in terms of quantitative and qualitative evaluations.
\end{itemize}

The source code of our implementation is available at \url{https://github.com/vc-bonn/nerfs-are-mirror-detectors}.

\section{Related Work}
\label{sec:related_work}

In the following, we provide a brief survey on the developments in the scope of neural scene representations and their extension towards specular surfaces.

\paragraph*{Neural scene representations.} Scene modeling and novel view synthesis based on neural scene representation and rendering techniques~\cite{tewari2020state,tewari2022advances} has gained a lot of attention in recent years due to the impressive results achieved by seminal work in this regard~\cite{lombardi2019neural,sitzmann2019scene,niemeyer2020differentiable,bi2020neural,bi2020deep,mildenhall2020nerf}.
In particular, neural radiance fields (NeRF)~\cite{mildenhall2020nerf} led to a breakthrough regarding the achievable reconstruction quality and photorealism in synthesized views of the (neural) scene models with a simple and effective formulation.
The impact of this approach is indicated by numerous extensions including the acceleration of the training~\cite{mueller2022instant,Mubarik2023HardwareAcc,wang2023neus2,sun2022direct,Cao2022rtNeRF,chen2022tensorf,fridovich2022plenoxels,Zhang2023Nerflets,kerbl20233d,deng2022depth,wang2022r2l,fang2022fast}, the acceleration of the rendering~\cite{reiser2021kilonerf,garbin2021fastnerf,chen2022mobilenerf,Yariv2023baked}, and the improvement of the scene representation and rendering quality in terms of reducing aliasing artifacts by replacing ray-based marching with an integration of 3D conical frustums~\cite{barron2021mipnerf,barron2022mipnerf360}, rendering fine details at very high-resolution~\cite{wang2022nerf,wang20224k,li2023uhdnerf} and combining scale-aware anti-aliased NeRFs and fast grid-based NeRF training based on multisampling and prefiltering techniques~\cite{barron2023zip}.
Further extensions include streamable representations~\cite{cho2022streamable,Takikawa2022VBNF}, the handling of  unbounded scenes~\cite{zhang2020nerf++,barron2022mipnerf360}, the handling of image collections taken under \emph{in-the-wild} conditions~\cite{martin2021nerf,chen2022hallucinated,Seong2022hdrplenoxels} or low-dynamic-range images with low or varying exposure~\cite{huang2022hdr,mildenhall2022nerf}, video inputs~\cite{li2021neural,xian2021space,du2021neural,peng2021neural,gao2021dynamic,li2022neural,Tancik2022blocknerf,Li2022streaming}, and the refinement or complete estimation of camera pose parameters for the input images~\cite{yen2021inerf,wang2021nerf,sucar2021imap,Chung-2022-Orbeez-SLAM,zhu2022nice,zhu2023nicer,rosinol2022nerf,zhang2022nerfusion,meng2021gnerf,lin2021barf,jeong2021self,Xia-2022-SiNeRF,Maggio-2023-Loc-NeRF,Bian-2022-NoPe-NeRF,Cheng20223LUNeRF,chen-2023-DBARF,Chen-2023-L2G-NeRF,Heo-2023-CamPos_MResHashEncoding,Sun2023NeRF-Loc,Liu2023NeRF-Loc} as well as the handling of deformable scenes~\cite{park2021nerfies,pumarola2021d,gafni2021dynamic,tretschk2021non,raj2021,noguchi2021neural,tseng2022cla,peng2021animatable,park2021hypernerf,chen2021animatable,liu2021neural,Jiang2022alignnerf,Li2022DynIBaR,fang2022fast} and large-scale scenarios~\cite{tancik2022block,turki2022mega,Mi2023switchnerf}.
In addition, several works focused on guiding the training and handling textureless regions by incorporating depth cues~\cite{wei2021nerfingmvs,deng2022depth,roessle2022dense,rematas2022urban,attal2021torf}.

Instead of employing a density-based shape representation, further techniques~\cite{zhang2021physg,wang2021neus,wang2023neus2,fan2023factored,liang2023envidr,wu2023nefii,ge2023ref} focused on scene representation in terms of implicit surfaces based on signed distance functions (SDF) to improve the estimation of object geometry and surface normals.

Despite the impressive progress offered by the aforementioned techniques, their limitations in modeling physical reflection within the rendering pipeline restrict their applicability to scenes featuring mirroring surfaces.
In such scenarios, these methods mistakenly interpret the mirror's reflection as an independent virtual scene. This misinterpretation creates an illusion of viewing the corresponding content through a window, ultimately resulting in an inaccurate reconstruction of the mirror's geometry and an incorrect, inconsistent appearance across multiple views.

\paragraph*{Neural scene representations for specular reflections.}
To allow handling surfaces with more general reflectance also including a specular reflectance component, several approaches focused on enriching NeRFs by an understanding of physical reflection~\cite{bi2020neural,boss2021nerd,srinivasan2021nerv,boss2022samurai,boss2021neural,jin2023tensoir,kuang2022neroic,munkberg2022extracting,zhang2021nerfactor,zhang2023nemf,zhao2022factorized}, which has been used for the decomposition of the radiance field into shape, material parameters (e.g., expressed through a Bidirectional Reflectance Distribution Function (BRDF)), and illumination characteristics in the scene to allow scene relighting.
Ref-NeRF~\cite{verbin2022refnerf} separates the light into diffuse and specular components and acquires knowledge about reflection by training a radiance field conditioned on the reflected view direction, where the observed radiance is reparametrized based on the local normal vector and its angle to the view direction, thereby simplifying the model and facilitating the sharing of structures across multiple views.
PhySG~\cite{zhang2021physg} focus on efficient approximate light transport in terms of representing specular BRDFs and environmental illumination using mixtures of spherical Gaussians and integrating the incoming light over the hemisphere of the surface.
Zhang \etal~\cite{zhang2022modeling} further model the indirect illumination and visibility of direct illumination to allow the recovery of interreflection- and shadow-free albedo.
Ref-NeuS~\cite{ge2023ref} focuses on the detection of anomalies in rendered images resulting from reflections and integrates a corresponding reflection score into the rendering loss for guidance. However, this method relies on masked input images that only depict the reflective surface parts and only approaches geometric surface reconstruction instead of novel view synthesis.
Further alternatives include directly tracing reflections in the ideal reflection direction based on the assumption of a low material roughness~\cite{liang2023envidr} or directly tracing reflections through path tracing evaluated using Monte-Carlo estimators~\cite{wu2023nefii}.
In addition, hybrid approaches such as volumetric microflake fields~\cite{zhang2023nemf} and microfacet fields~\cite{mai2023neural} allow the combination of ray marching and path tracing with an importance sampling accounting for the distribution of microstructures.

Most closely related to our work are techniques that explicitly focus on the modeling of mirror reflections within the scene and the reconstruction of mirroring surfaces.
Mirror-NeRF~\cite{zeng2023mirror} relies on a single radiance field and tracing the rays into the direction of ideal reflection after hitting a mirror. Respective normal directions and reflection probabilities used for reflecting the rays are learned in the volumetric neural field, similar as in the microflake/microfacet fields~\cite{zhang2023nemf,mai2023neural}.
Casting reflection rays for the points along the camera rays for standard NeRF points into the scene instead of querying view-dependent appearance for points along rays has also been followed by others~\cite{verbin2024nerf,ma2023specnerf}.
Furthermore, advanced view-dependent appearance encoding of NeRFs~\cite{wu2024neural} have been proposed, where the involved MLP takes features obtained for encoding far-field reflections into a cubemap and near-field interreflections as learnable parameters into a volume, and predicts specular color.
However, such neural directional encoding is affected by the sensitivity to the quality of the surface normal.
NeRFReN~\cite{guo2022nerfren} decomposes the scene into a transmitted component and a reflected component that are separately modeled based on respective neural radiance fields, thereby involving geometric priors and an advanced training strategy to improve the robustness of the decomposition. For view synthesis, the results for the transmitted and reflected fields are blended.
MS-NeRF~\cite{yin2023msnerf} represent the scene based on multiple parallel neural feature fields~\cite{niemeyer2021giraffe} to improve the neural network's awareness regarding the presence of reflective and refractive objects.
Others~\cite{meng2024mirror,liu2024mirrorgaussian} used a combination of a standard Gaussian splatting representation and a Gaussian splatting representation for planar, mirroring surfaces which relies on the reflection of the former across the mirror plane.
These components are then combined based on a dual-rendering strategy.
However, both approaches rely on providing segmentations of the mirroring surface parts in the input images and are restricted to planar, mirroring objects.
Furthermore, Ye \etal~\cite{ye20243d} use deferred shading to get per-pixel reflection gradients, thereby allowing the propagation of normal estimates over reflective objects.
However, the use of deferred shading results in the limitation of only being suitable for the handling of at most one layer of reflective materials per pixel and convergence has been shown to be less efficient for concave scene parts.
TraM-NeRF~\cite{holland2023tramnerf} relies on a transmittance-aware formulation of the rendering equation to explicitly model reflected radiance at mirror-like surfaces and improving the efficiency in comparison to Monte-Carlo estimation approaches based on efficient importance sampling and transmittance computation.
In contrast, to the aforementioned methods, our approach allows both the automated localization and reconstruction of mirroring surfaces within larger scenes in a NeRF-based framework.

\section{Background}
\label{sec:method}

\subsection{Neural Radiance Fields}
\label{sec:nerfs}
Given a set of $N$ training images and corresponding camera transformations, we use an initial training phase based on the standard NeRF approach~\cite{mildenhall2020nerf}.
A small feed-forward network is used to predict a radiance ${c(x)\in[0,1]^3}$ and volume density ${\sigma(x)\in[0,1]}$ for a given spatial location ${x\in\mathbb{R}^3}$, which can then be used to render the observed color by integrating along viewing rays ${r(t) = o + t \, d}$ in the volume, where ${o,d\in\mathbb{R}^3}$ denotes the origin and direction of the ray.
This integral can be estimated by sampling ${K\in\mathbb{N}}$ positions ${t_1, \dots, t_K\in\mathbb{R}}$ along the ray~\cite{max1995optical}:
\begin{equation}
    C(r) = \sum_{k=1}^K T_k \, \alpha_k \, c_k,
\label{eq:nerf_color}
\end{equation}
where ${\alpha_k = 1 - e^{-\sigma_k \delta_k}}$ is the probability that the ray will hit a particle at position $t_k$, ${T_k = \exp(-\sum_{j=1}^{k-1} \sigma_j \, \delta_j)}$ is the transmittance along the ray up to $t_k$, while ${c_k = c(r(t_k))}$ and ${\sigma_k=\sigma(r(t_k))}$ are the radiance and density at $t_k$ respectively and ${\delta_k = t_{k+1} - t_k}$ is the length of the integrated segment.
The positions $t_k$ are generated in a hierarchical fashion, where an initial stratified sampling is refined based on the density distribution along each ray~\cite{mildenhall2020nerf}.
Based on this formulation, the optimization loss is then defined as the mean squared error between the rendered color $C(r)$ and the corresponding color from the input image $C^*(r)$ for a batch of camera rays $R$:
\begin{equation}
    \mathcal{L}_\text{I} =\frac{1}{\abs{R}} \sum_{r\in R} \norm{C(r)-C^*(r)}_2^2
    \label{eq:image_loss}
\end{equation}
Note that we also employ the integrated positional encoding proposed by Barron \etal~\cite{barron2021mipnerf} for the input of the network which, for the sake of brevity, has been omitted here.

\subsection{Plausible Depths during Training}
\label{sec:plausible_depths}

While the NeRF optimization process is able to reproduce the ground-truth image data with high quality, the learned geometry often remains implausible, especially in textureless or distant regions of the scene.
To improve the depth predictions during training, we use a depth reprojection loss similar to Truong \etal~\cite{truong2023sparf}.

We define the depth $D(r)$ of ray $r$ of camera $i$ as the expected absorption position along the ray, \ie
\begin{equation}
     D(r)=\sum_{k=1}^K T_k \, \alpha_k \, t_k.
\label{eq:expected_ray_termination}
\end{equation}
Let ${j \not= i\in\mathbb{N}}$ be a camera index selected uniformly at random from the training dataset and $r_j$ be the ray from the origin of camera $j$ through the 3D absorption position ${o + D(r)\, d}$ of $r$ as seen from camera $i$.
We define ${z_j(D(r), r))}$ as the depth of that absorption position in camera $j$, such that the approximate reprojection loss is given as
\begin{equation}
    \mathcal{L}_\text{D}(R)=\frac{1}{\abs{R}}\sum_{r\in R} w_{ij} \abs{D(r_j) - z_j(D(r),r)}^2,
\label{eq:pc_loss}
\end{equation}

where ${w_{ij}=\frac{1}{w_\text{max}} \, \abs{\phi_{ij}} \, \norm{o_i - o_j}_2\in[0,1]}$ weights the color error by both the Euclidean distance of the camera origins and the angle $\phi_{ij}$ between the optical axes of the cameras.
${w_\text{max}=\max_{i,j} w_{ij}}$ normalizes the weighting.
This choice of $w_{ij}$ reduces the effect of occlusions on the measurement, assuming more dissimilar cameras are less likely to observe the same surface point along intersecting rays.

\section{NeRFs with Reflections via Mirror Detection}

\begin{figure*}
    \centering
    \includegraphics[width=\linewidth,keepaspectratio]{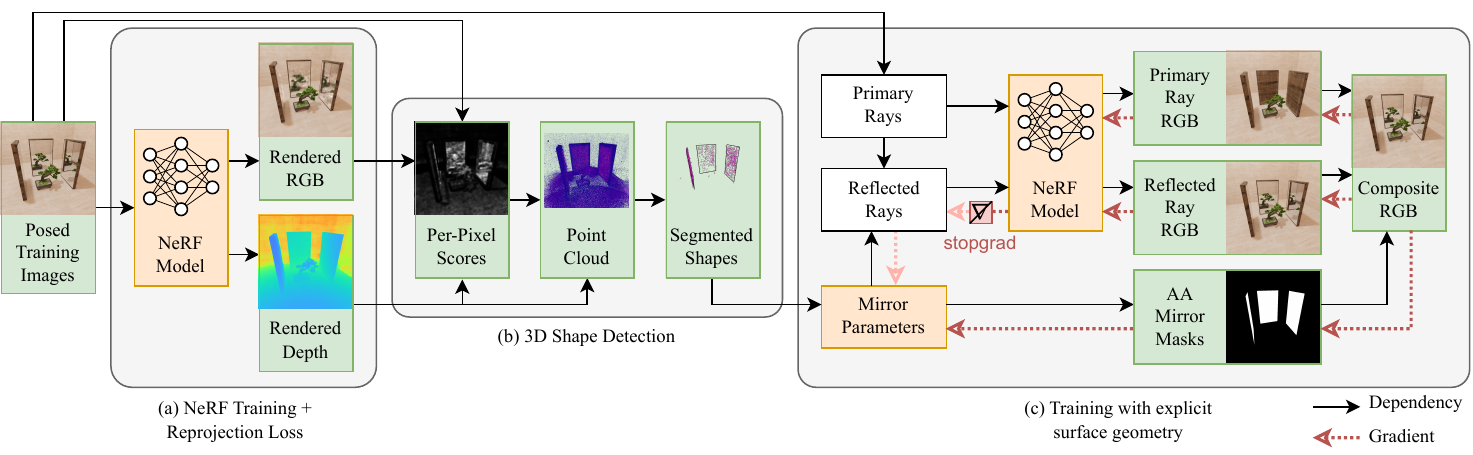}
    \caption{Overview of our pipeline.
    (a) A collection of training images is first used to train a standard NeRF model with an additional depth reprojection loss (\cref{eq:pc_loss}).
    (b) Per-pixel scores are then computed using SSIM and depth variance values.
    Highly scoring pixels are unprojected into 3D, and the resulting point cloud is segmented into primitive shapes.
    (c) Finally, a modified rendering pipeline is employed to jointly optimize NeRF and mirror parameters by blending primary and reflected images together based on antialiased mirror masks that are generated in a differentiable manner.}
    \label{fig:pipeline}
    \vspace{-12pt}
\end{figure*}

Given a set of posed images like in the standard NeRF setting, we aim at training a NeRF-based scene representation that is not only able to produce novel views of a scene containing mirroring surfaces, but also represents the scene geometry in a consistent manner.
For this purpose, we propose a pipeline, depicted in \cref{fig:pipeline}, that consists of three steps:
1) In the initial step, we train a neural radiance field to get insights on multi-view-consistent (\ie, non-mirroring) scene structures.
2) Next, we use the resulting initial geometry estimate of the trained field to analyze surface points with respect to the rendering error of the initial training.
This is based on the observation that standard NeRF approaches struggle to achieve high reconstruction quality in the presence of highly reflective surfaces, but only in image regions that contain these surfaces \cite{holland2023tramnerf}.
An example of these artifacts can be seen in \cref{fig:score_values_rendered_rgb}.
This allows us to obtain a set of mirror candidate points by analyzing the rendering error and unproject the highly erroneous pixels into 3D.
The resulting point cloud is then segmented into a set of 3D primitives.
3) Finally, these primitives are used in an extended neural radiance field that models ray reflections at the surfaces explicitly and is differentiable with respect to the parameters of the primitives.
This way, we are able to jointly optimize the radiance field and the primitive.

In the following, we will explain each component of the pipeline in detail.
Hyperparameter choices and implementation details can be found in the supplementary material.

\subsection{Structural Similarity Analysis}
\label{sec:photometric_consistency_analysis}
\begin{figure}
    \centering
    \begin{subfigure}[b]{0.326\linewidth}
        \includegraphics[width=\linewidth]{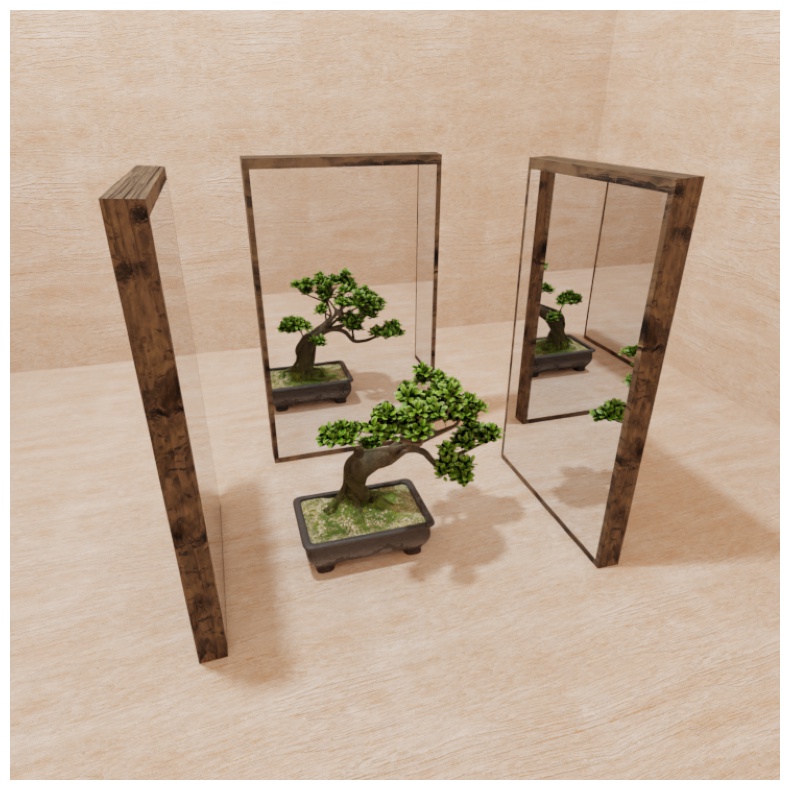}
        \caption{GT $C^*(r)$}
        \label{fig:score_values_gt}
    \end{subfigure}
    \hfill
    \begin{subfigure}[b]{0.326\linewidth}
        \includegraphics[width=\linewidth]{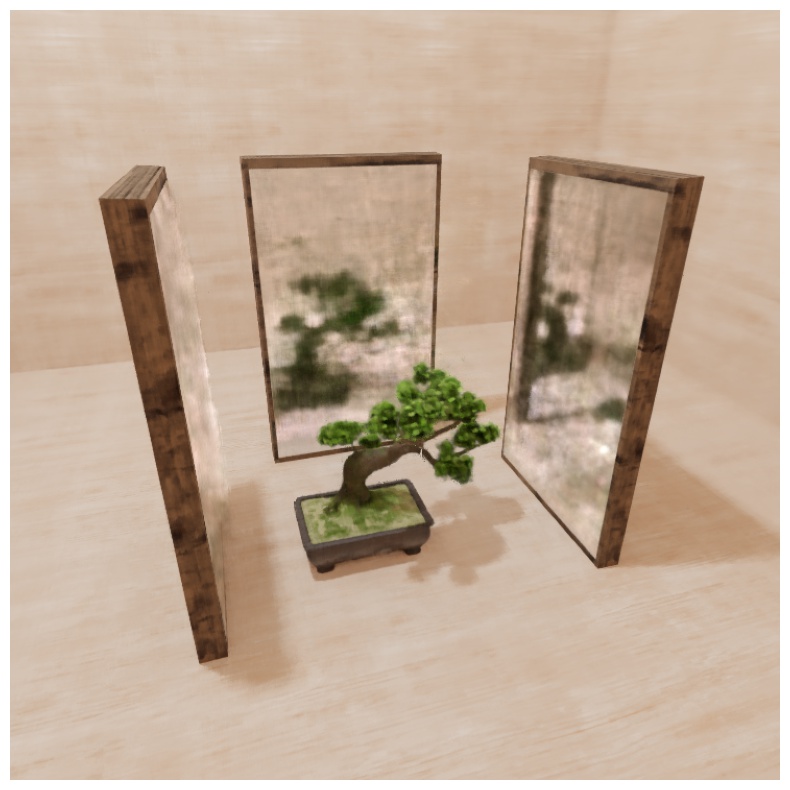}
        \caption{Rend. RGB $C(r)$}
        \label{fig:score_values_rendered_rgb}
    \end{subfigure}
    \hfill
    \begin{subfigure}[b]{0.326\linewidth}
        \includegraphics[width=\linewidth]{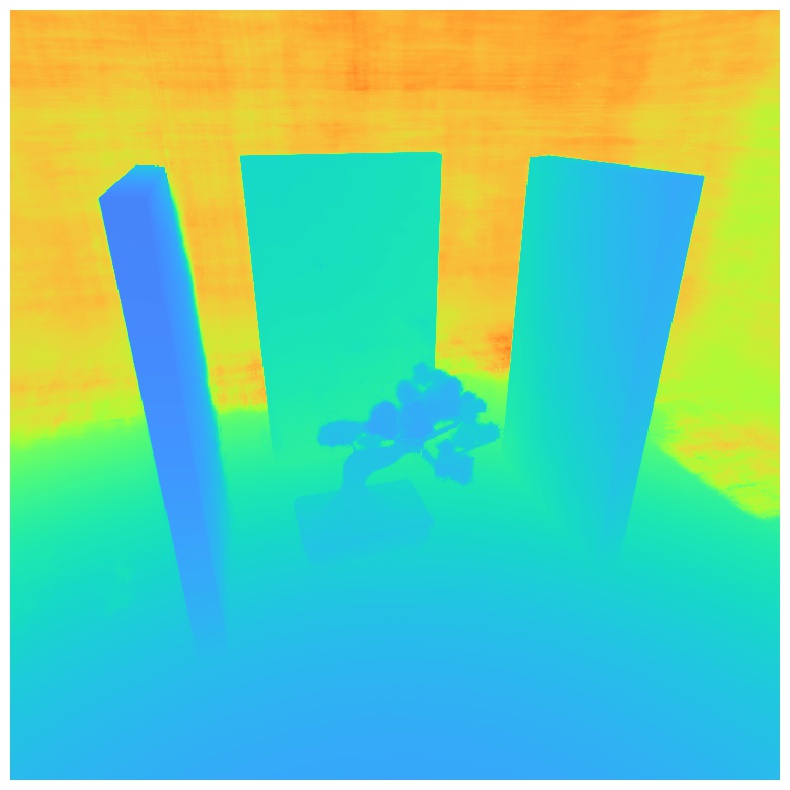}
        \caption{Rend. Depth $D(r)$}
        \label{fig:score_values_rendered_depth}
    \end{subfigure}
    \\
    \begin{subfigure}[b]{0.326\linewidth}
        \includegraphics[width=\linewidth]{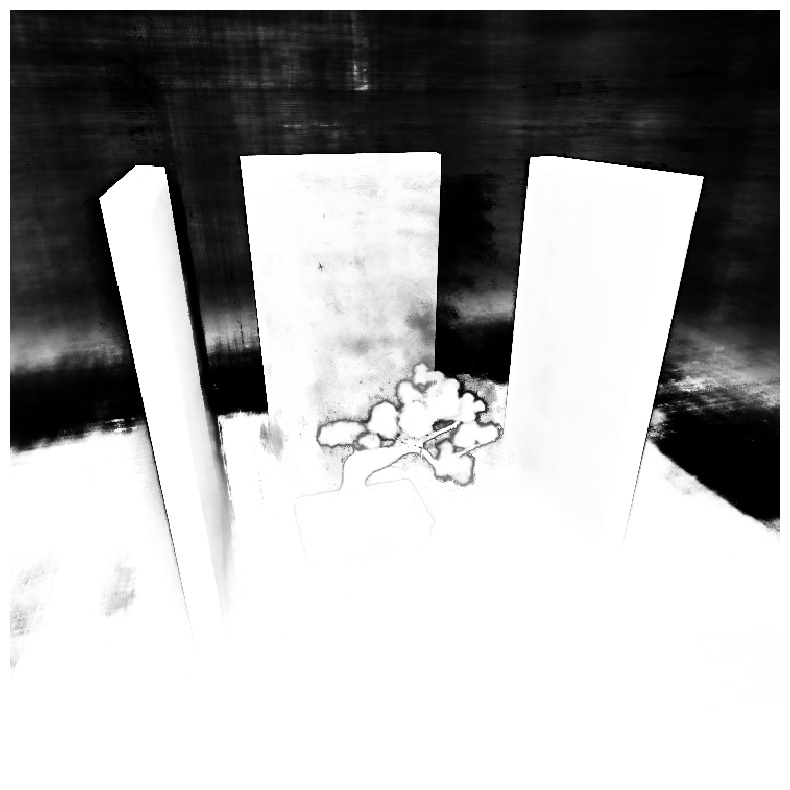}
        \caption{Depth Var. $e^{-c \, V(r)}$}
        \label{fig:score_values_depth_var}
    \end{subfigure}
    \hfill
    \begin{subfigure}[b]{0.326\linewidth}
        \includegraphics[width=\linewidth]{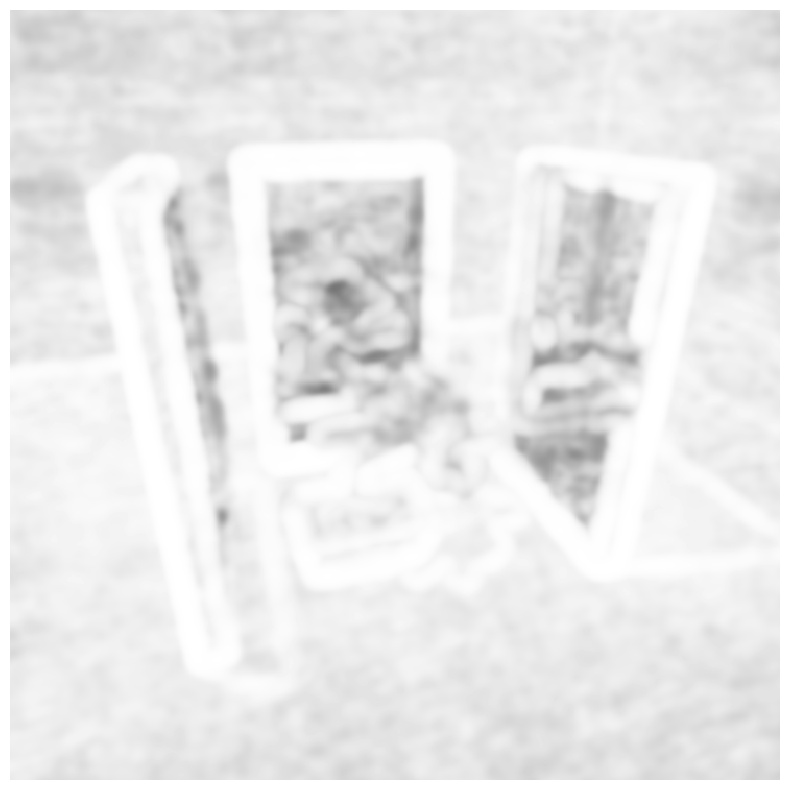}
        \caption{$\text{SSIM}(r)$}
        \label{fig:score_values_ssim}
    \end{subfigure}
    \hfill
    \begin{subfigure}[b]{0.326\linewidth}
        \includegraphics[width=\linewidth]{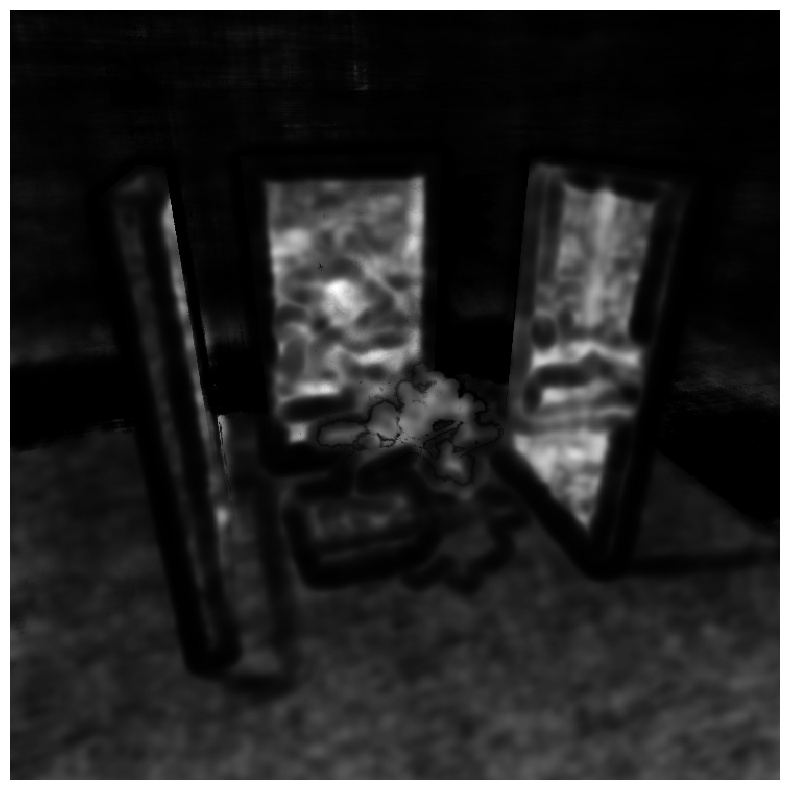}
        \caption{Final Score $s(r)$}
        \label{fig:score_values_final_score}
    \end{subfigure}
    \caption{Examples for the intermediate values used to generate the final score $s(r)$.}
    \label{fig:score_values}
    \vspace{-12pt}
\end{figure}
After the initial training of the radiance field has been completed, we compare the rendered colors of the training images to the ground-truth using SSIM and derive a score $s(r)$ that quantifies if the pixel corresponding to ray $r$ exhibits artifacts that indicate poor convergence of the initial training.
An example for each of the values involved in the computation is given in \cref{fig:score_values}.
More specifically, we look at the similarity score of a local window around the pixel corresponding to ray $r$ given by
\begin{equation}
    \text{SSIM}(r) = \frac{(2\,\mu\,\mu^*+c_1)(2\tilde{\sigma}+c_2)}{\left(\mu^2+(\mu^*)^2+c_1\right)\left(\sigma^2+(\sigma^*)^2+c_2\right)}
\end{equation}
where $\mu, \mu^*$ are means, $\sigma, \sigma^*$ are variances and $\tilde{\sigma}$ is the covariance of pixel intensities of the neighborhood of the pixel corresponding to ray $r$ in the rendered and ground-truth image respectively. ${c_1, c_2 \in \mathbb{R}}$ are constants chosen as proposed by Wang \etal~\cite{wang2004ssim}.
Using a metric that considers a pixel neighborhood instead of a single measurement is important here, as highly reflective surfaces exhibit reduced sharpness in the renderings, which is difficult to quantify with, \eg, the squared error used for training.
To reduce the influence of regions with uncertain depth estimates, we compute the variance of the expected absorption position
\begin{equation}
    V(r)=\sum_{k=1}^K T_k \, \alpha_k (t_k - D(r))^2
\end{equation}
and use it to compute re-weighted SSIM scores,
\begin{equation}
    %s(p)=\tanh\left(\frac{1-\text{SSIM}(p)}{c\sqrt{V_i(r)}}\right)
    s(r)=\frac{1 - \text{SSIM}(r)}{2}\, e^{-c \, V(r)},
\end{equation}
where ${c\in\mathbb{R}}$ is used to control the slope of the mapping.
After this, a high score indicates regions with low structural similarity, but high certainty in the depth estimate.

\subsection{Primitive Shape Fitting}
\label{sec:primitive_shape_fitting}
Using the scores obtained in the previous step, we generate a point cloud from pixels that have a score $s(t)$ above some threshold ${S\in[0,1]}$, using the depth values computed with \cref{eq:expected_ray_termination}.
To acquire additional information for the fitting step, we then estimate normals for the point cloud where we gather neighbors using a hybrid strategy based on radius and a $k$-nearest neighbor search, which helps to handle regions of varying density.
As each point in the point cloud is generated by a single camera, we can use the camera position to solve the orientation ambiguity.
Given a type of primitive shape and the number of shapes $k$ to detect, we now segment the point cloud into separate regions using $k$-means clustering \cite{lloyd1982least}.
For each of the regions, we found that using RANSAC \cite{fischler1981random} to fit the primitive shape to the point suffices to yield a plausible initial estimate of the surface.
More specifically, we first compute a best fit of the unbounded shape onto which the primitive lies and then compute an oriented bounding box of the projected inliers after filtering out low-density areas using a bitmap.
The remaining inliers are then evaluated using three metrics: 1) Inlier ratio, 2) average shortest distance to the unbounded shape, and 3) similarity of the normal, which the model predicts at the projected position, to the estimated point cloud normal.

\subsection{Training Scenes with Reflections}
\label{sec:method_nerf_reflections}

With a first estimate of the primitive shapes, given as a set of parameters $\theta$, we are now equipped to run a modified version of the NeRF training that considers explicit mirror geometry, following TraM-NeRF~\cite{holland2023tramnerf}.
More specifically, in each iteration we check the rays in the current batch for intersections with any of the primitive shapes recursively until a predefined bounce limit is reached, reflecting the camera ray at the surface if an intersection occurred.
As the intersection computation is implemented in a differentiable manner with respect to the geometry parameters, we could continue fine-tuning the initial shape prediction using the image loss during the training.
However, we found that the gradients produced by the NeRF rendering process do not provide a stable signal for optimization of the mirror parameters.
To alleviate this issue, we instead render antialiased mirror masks in a differentiable manner \cite{Laine2020diffrast} and, for pixel frustums that are only partially intersecting the surface, we render both primary and reflected rays and blend them together proportional to the surface area intersected by the frustum.
This allows us to propagate the gradient of the image loss back through the mask generation instead, skipping the propagation through \cref{eq:nerf_color} altogether.
\begin{figure}
    \centering
    \includegraphics[width=\linewidth]{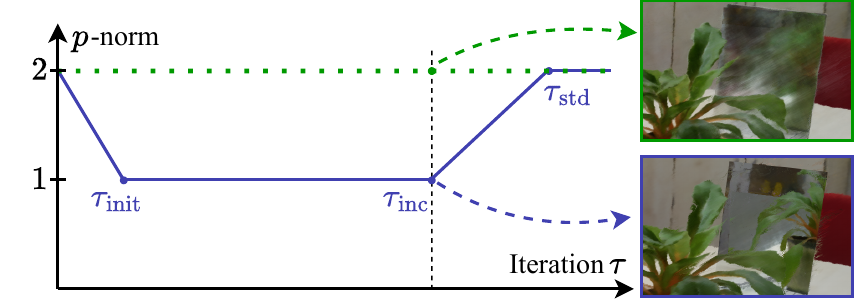}
    \caption{The proposed $p$-norm schedule (solid blue line) compared to the usual $L_2$ optimization (dotted green line). The images show a zoom in to a mirror region after $\tau_\text{inc}$ iterations.}
    \label{fig:pnorm}
    \vspace{-12pt}
\end{figure}
Especially for the more challenging scenario of real-world scenes, we additionally found that the optimization requires a loss that favors sharp features, as otherwise the optimization will quickly fall back to an optimum without explicit mirrors, as can be seen in \cref{fig:pnorm}.
While an $L_1$ loss is known to favor this property, we observed that the training process either fails to leave a low-density optimum present in the beginning of the training or exhibits regions of locally constant color in the final reconstruction if \cref{eq:image_loss} is replaced with an $L_1$ loss.
To address both issues, we propose a scheduling scheme that modulates the order of the norm during the course of the training.
We consider the order as a function of the iterations, ${p(\tau)\in [1, 2]}$, and generalize \cref{eq:image_loss} to
\begin{equation}
    \mathcal{L}_\text{I}'=\frac{1}{\abs{R}}\sum_{r\in R} \norm{C(r)-C^*(r)}_{p(\tau)}^{p(\tau)}.
\end{equation}
For $p$, we choose a piecewise linear function that starts out with ${p=2}$, reaches a plateau of ${p=1}$ after $\tau_\text{init}$ iterations to optimize the mirror positions while encouraging sharp edges in the resulting images and increase back to $p=2$ after $\tau_\text{inc}$ iterations, reaching ${p=2}$ after $\tau_\text{std}$ iterations to remove the aforementioned artifacts.
The function, together with examples of the mirror regions after $\tau_\text{inc}$ iterations, is visualized in \cref{fig:pnorm}.

\section{Results}
\label{sec:results}

\begin{table*}[ht]
    \centering
    \fontsize{8pt}{8pt}\selectfont
    \setlength{\tabcolsep}{1.8pt} % 6pt is default
    \begin{tabular}{lccccccc}
        \toprule
        & & \multicolumn{3}{c}{\textbf{Synthetic}} & \multicolumn{3}{c}{\textbf{Real World}} \\
        \cmidrule(lr){3-5} \cmidrule(lr){6-8}
        \textbf{Full Images} & Prior & PSNR $\uparrow$ & SSIM $\uparrow$ & LPIPS $\downarrow$ & PSNR $\uparrow$ & SSIM $\uparrow$ & LPIPS $\downarrow$ \\
        \midrule
NeRF \cite{mildenhall2020nerf} & \xmark
& $26.53 \pm 2.86$
& $0.819 \pm 0.035$
& $0.370 \pm 0.057$
& $24.32 \pm 0.79$
& $0.799 \pm 0.001$
& $0.443 \pm 0.013$ \\
Mip-NeRF360 \cite{barron2022mipnerf360} & \xmark
& $26.05 \pm 5.05$
& $0.775 \pm 0.122$
& $0.413 \pm 0.058$
& $25.88 \pm 1.68$
& \colorbox{best}{$0.862 \pm 0.025$}
& \colorbox{best}{$0.310 \pm 0.032$} \\
Ref-NeRF \cite{verbin2022refnerf} & \xmark
& $25.67 \pm 4.87$
& $0.784 \pm 0.102$
& $0.418 \pm 0.045$
& $25.30 \pm 1.64$
& $0.821 \pm 0.014$
& $0.410 \pm 0.024$ \\
MS-NeRF \cite{yin2023msnerf} & \xmark
& $29.13 \pm 4.55$
& \colorbox{second}{$0.836 \pm 0.060$}
& $0.377 \pm 0.073$
& $25.51 \pm 3.47$
& $0.805 \pm 0.085$
& $0.429 \pm 0.124$ \\
\midrule
Mirror-NeRF \cite{zeng2023mirror} & \cmark
& $21.49 \pm 3.96$
& $0.712 \pm 0.075$
& $0.526 \pm 0.088$
& $16.37 \pm 5.94$
& $0.577 \pm 0.224$
& $0.587 \pm 0.135$ \\
TraM-NeRF \cite{holland2023tramnerf} & \cmark
& \colorbox{best}{$31.46 \pm 3.31$}
& \colorbox{best}{$0.875 \pm 0.026$}
& \colorbox{second}{$0.285 \pm 0.047$}
& \colorbox{best}{$26.66 \pm 0.56$}
& \colorbox{second}{$0.842 \pm 0.015$}
& $0.394 \pm 0.006$ \\
\midrule
Ours & \xmark
& \colorbox{second}{$30.30 \pm 4.26$}
& \colorbox{best}{$0.875 \pm 0.047$}
& \colorbox{best}{$0.263 \pm 0.068$}
& \colorbox{second}{$26.20 \pm 0.54$}
& $0.838 \pm 0.013$
& \colorbox{second}{$0.392 \pm 0.002$} \\
        \midrule
        \midrule
        \textbf{Mirror Regions} & & & & & & \\
        \midrule
        NeRF \cite{mildenhall2020nerf} & \xmark
& $28.55 \pm 2.09$
& $0.945 \pm 0.014$
& $0.070 \pm 0.021$
& $28.47 \pm 3.46$
& $0.944 \pm 0.040$
& $0.082 \pm 0.062$ \\
Mip-NeRF360 \cite{barron2022mipnerf360} & \xmark
& $28.96 \pm 2.48$
& $0.946 \pm 0.013$
& $0.069 \pm 0.020$
& $28.86 \pm 3.07$
& $0.946 \pm 0.040$
& $0.075 \pm 0.060$ \\
Ref-NeRF \cite{verbin2022refnerf} &\xmark
& $29.01 \pm 2.67$
& $0.950 \pm 0.014$
& $0.070 \pm 0.021$
& $28.85 \pm 4.38$
& $0.941 \pm 0.048$
& $0.083 \pm 0.070$ \\
MS-NeRF \cite{yin2023msnerf} & \xmark
& $33.48 \pm 4.49$
& $0.965 \pm 0.018$
& $0.060 \pm 0.026$
& $31.86 \pm 2.78$
& $0.956 \pm 0.032$
& $0.077 \pm 0.064$ \\
\midrule
Mirror-NeRF \cite{zeng2023mirror} & \cmark
& $25.82 \pm 3.37$
& $0.933 \pm 0.021$
& $0.081 \pm 0.027$
& $24.29 \pm 0.51$
& $0.931 \pm 0.027$
& $0.090 \pm 0.056$ \\
TraM-NeRF \cite{holland2023tramnerf} & \cmark
& \colorbox{best}{$38.23 \pm 2.74$}
& \colorbox{best}{$0.982 \pm 0.007$}
& \colorbox{best}{$0.034 \pm 0.018$}
& \colorbox{best}{$34.00 \pm 3.00$}
& \colorbox{best}{$0.966 \pm 0.019$}
& \colorbox{best}{$0.065 \pm 0.049$} \\
\midrule
Ours & \xmark
& \colorbox{second}{$35.94 \pm 4.29$}
& \colorbox{second}{$0.978 \pm 0.015$}
& \colorbox{second}{$0.037 \pm 0.025$}
& \colorbox{second}{$32.20 \pm 2.38$}
& \colorbox{second}{$0.960 \pm 0.021$}
& \colorbox{second}{$0.070 \pm 0.051$} \\
        \bottomrule
    \end{tabular}
    \caption{Quantitative comparison of our approach against NeRF baselines \cite{mildenhall2020nerf,barron2022mipnerf360,verbin2022refnerf,yin2023msnerf} and recent works using mirror priors \cite{zeng2023mirror,holland2023tramnerf} on synthetic and real world multi-mirror scenes. Metrics are averaged over all test images and across all scenes. \colorbox{best}{Best} and \colorbox{second}{second-best} results are highlighted.}
    \label{tab:results}
\end{table*}
We evaluate our method on a subset of the dataset from \cite{holland2023tramnerf}, removing scenes that are either front-facing or containing specular surfaces with non-zero roughness.
This amounts to a total of seven synthetic and two real-world scenes with one or multiple planar mirrors.
The reconstructions are evaluated on a separate test set of images not seen during training using standard metrics, mean-squared error (MSE), structural similarity index measure (SSIM) \cite{wang2004ssim}, and learned perceptual image patch similarity (LPIPS) \cite{zhang2018lpips}.
We compare the results against baseline methods \cite{mildenhall2020nerf,barron2022mipnerf360} and methods that handle scenes with reflective surfaces explicitly \cite{verbin2022refnerf,yin2023msnerf}, and also report the scores to methods where the surfaces are explicitly annotated \cite{zeng2023mirror,holland2023tramnerf} to show how close our automatic detection can get to high-quality manual annotations.
The latter methods are marked with ``Prior'' in \cref{tab:results}.

Additionally, we report the reconstruction quality in mirror regions to show the distinct quality improvement on the mirror surfaces.
To select these regions, we use annotated masks provided by Holland \etal \cite{holland2023tramnerf} and define the metrics in the same way.
For a fair evaluation, we use the configuration parameters provided by the respective authors and matched the number of model parameters and iterations to previous evaluations~\cite{holland2023tramnerf}.

\begin{figure}
    \centering
    \captionsetup[subfigure]{justification=centering}
    \begin{subfigure}[t]{0.32\linewidth}
        \imagewithzoom{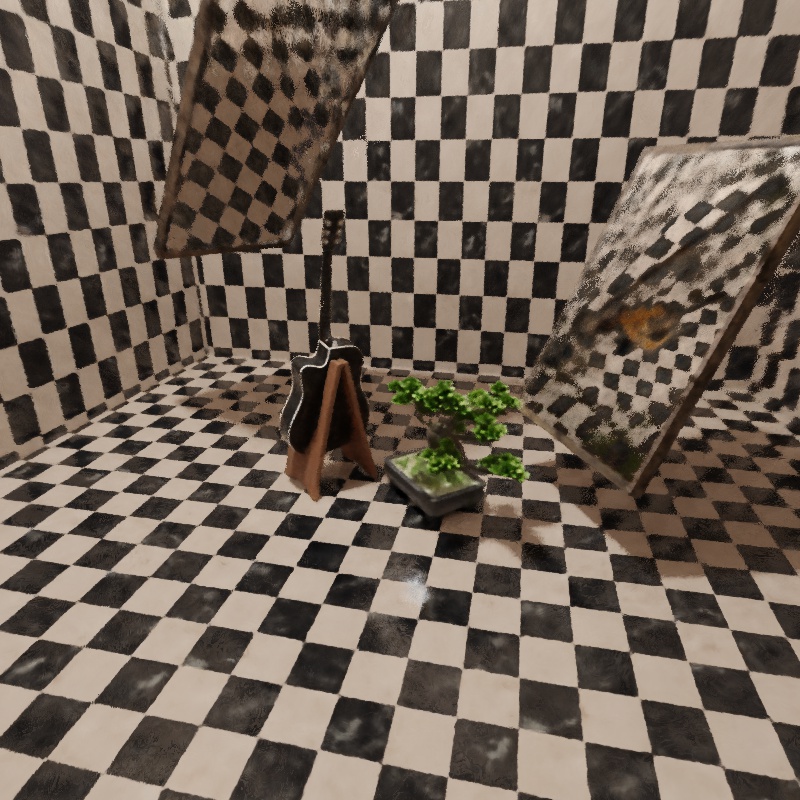}{0.85}{0.25}{-0.85}{-0.85}{0.8cm}{3}
    \end{subfigure}
    \hfill
    \begin{subfigure}[t]{0.32\linewidth}
        \imagewithzoom{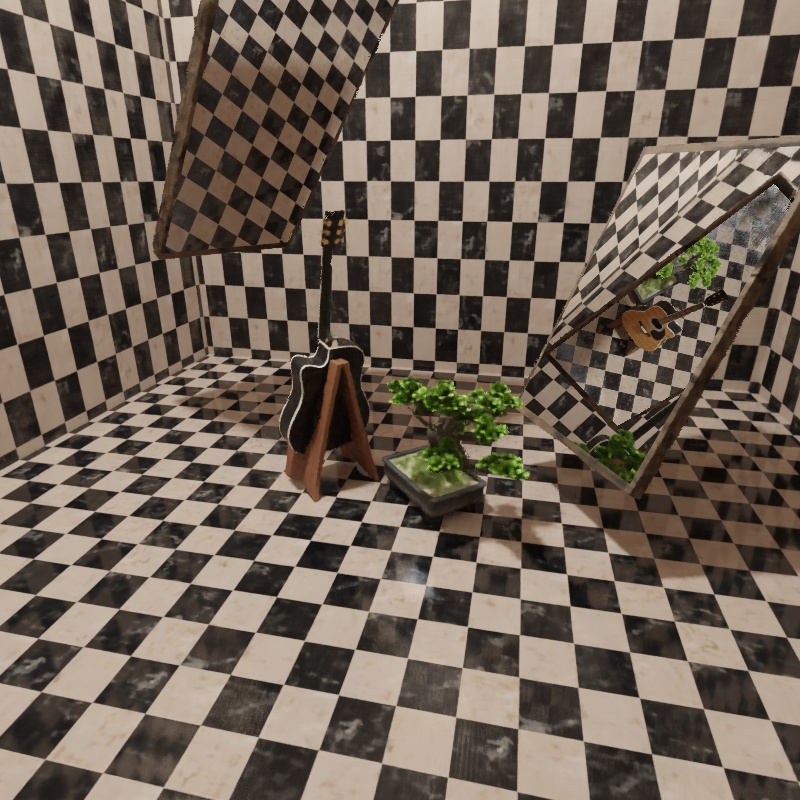}{0.85}{0.25}{-0.85}{-0.85}{0.8cm}{3}
    \end{subfigure}
    \hfill
    \begin{subfigure}[t]{0.32\linewidth}
        \imagewithzoom{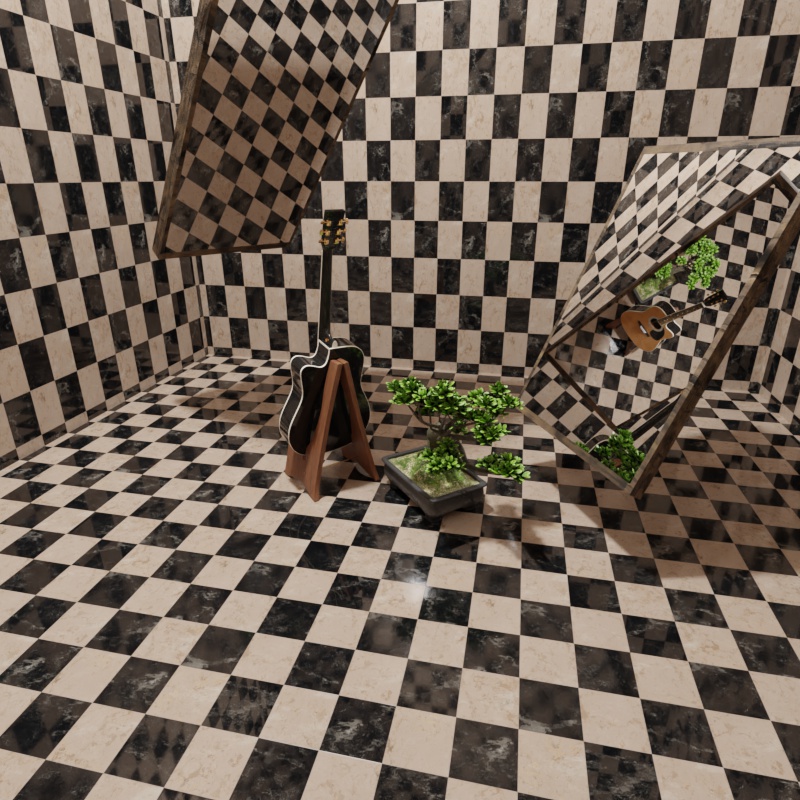}{0.85}{0.25}{-0.85}{-0.85}{0.8cm}{3}
    \end{subfigure}
    \\
    % %%%%%%%%%%%%
    \begin{subfigure}[t]{0.32\linewidth}
        \imagewithzoom{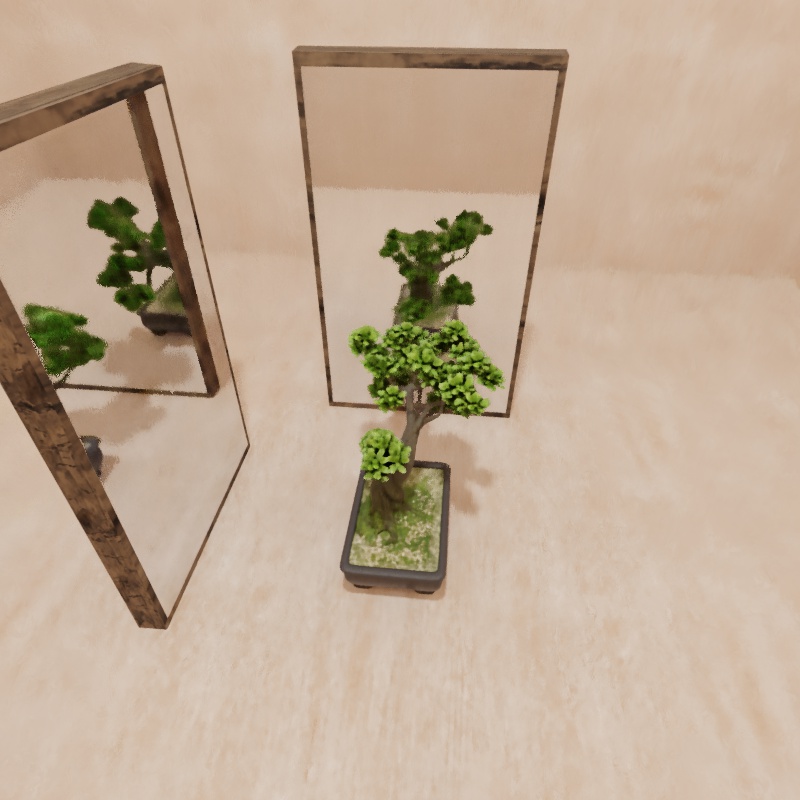}{-0.9}{0.4}{0.85}{-0.85}{0.8cm}{3}
        \caption*{Best Baseline \\ MS-NeRF \cite{yin2023msnerf}}
    \end{subfigure}
    \hfill
    \begin{subfigure}[t]{0.32\linewidth}
        \imagewithzoom{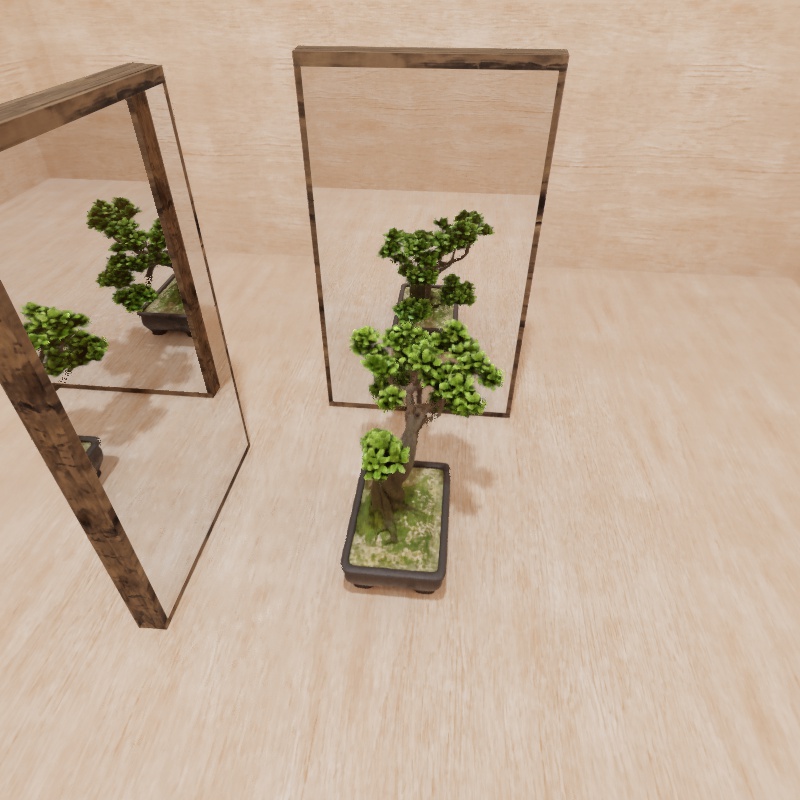}{-0.9}{0.4}{0.85}{-0.85}{0.8cm}{3}
        \caption*{Ours}
    \end{subfigure}
    \hfill
    \begin{subfigure}[t]{0.32\linewidth}
        \imagewithzoom{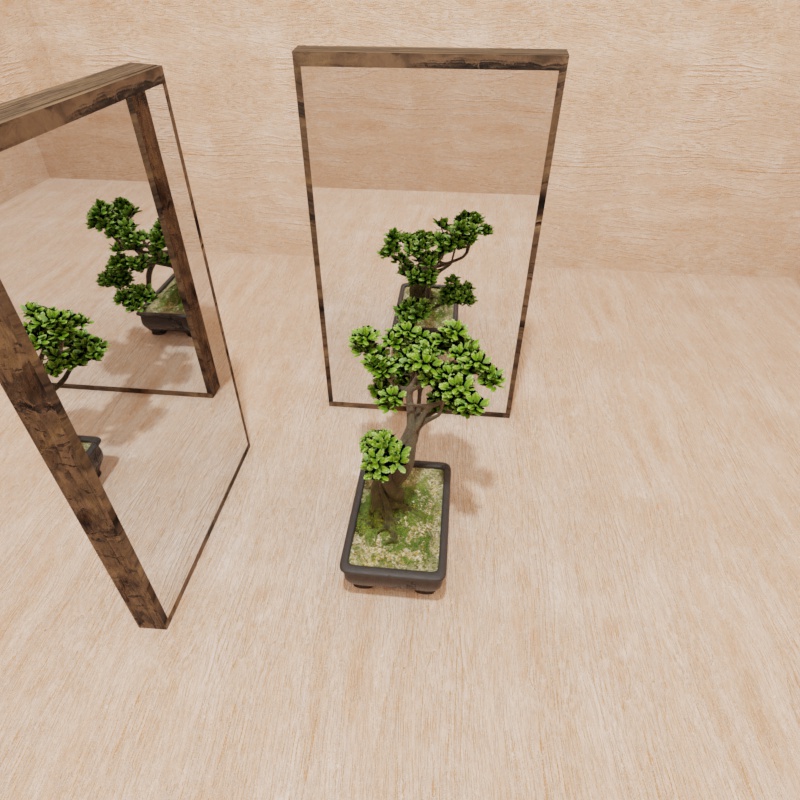}{-0.9}{0.4}{0.85}{-0.85}{0.8cm}{3}
        \caption*{Ground Truth}
    \end{subfigure}
    \caption{Results on the test set of two synthetic scenes compared to the best baseline method according to \cref{tab:results}.}
    \label{fig:results_qualitative_synthetic}
\end{figure}

\begin{figure}
\captionsetup[subfigure]{justification=centering}
    \begin{subfigure}[t]{0.32\linewidth}
        \imagewithzoom{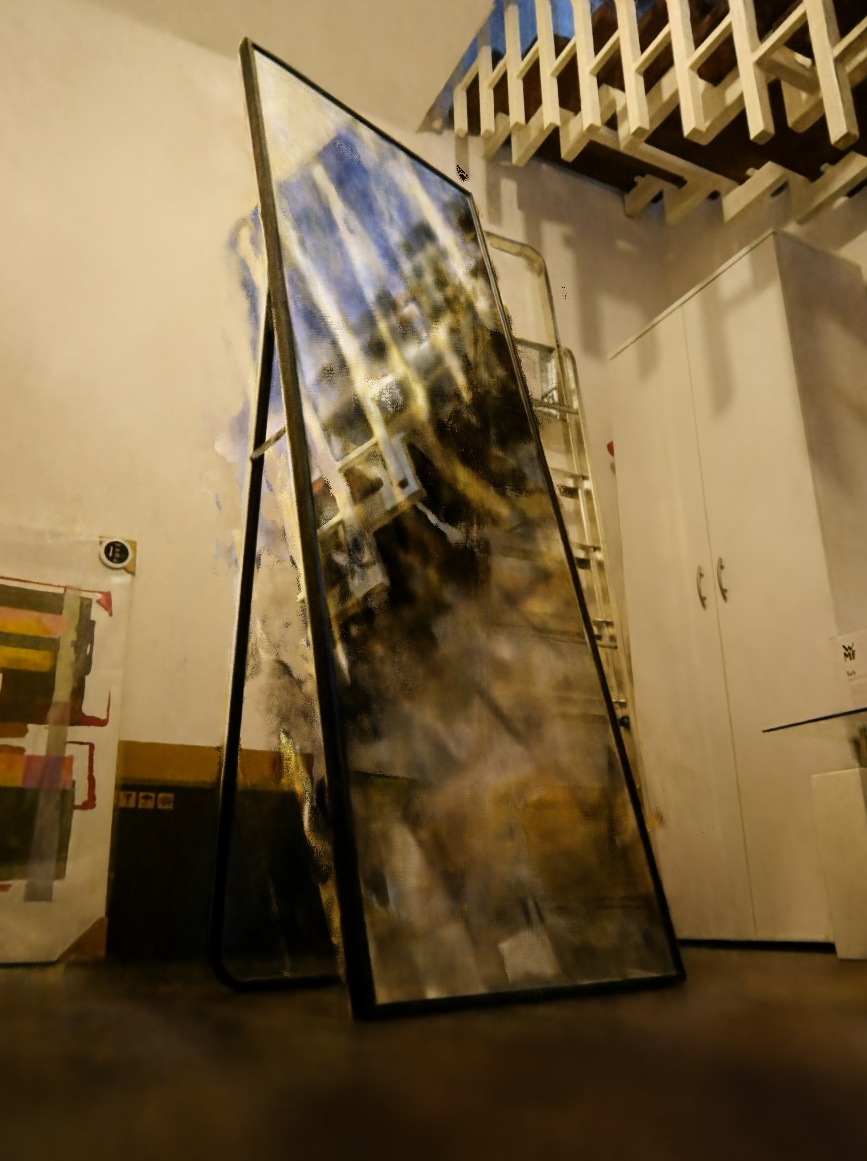}{0}{0.9}{-0.85}{-1.3}{0.8cm}{2}
    \end{subfigure}
    \hfill
    \begin{subfigure}[t]{0.32\linewidth}
        \imagewithzoom{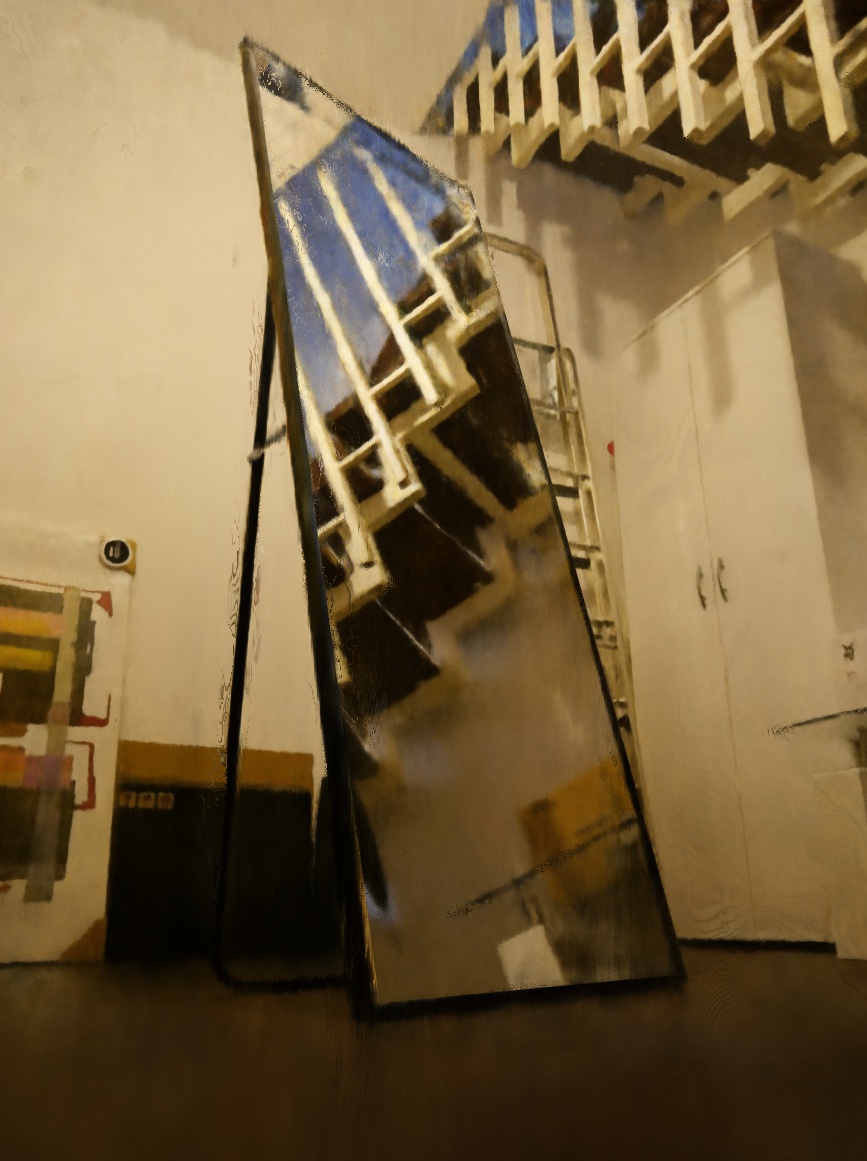}{0}{0.9}{-0.85}{-1.3}{0.8cm}{2}
    \end{subfigure}
    \hfill
    \begin{subfigure}[t]{0.32\linewidth}
        \imagewithzoom{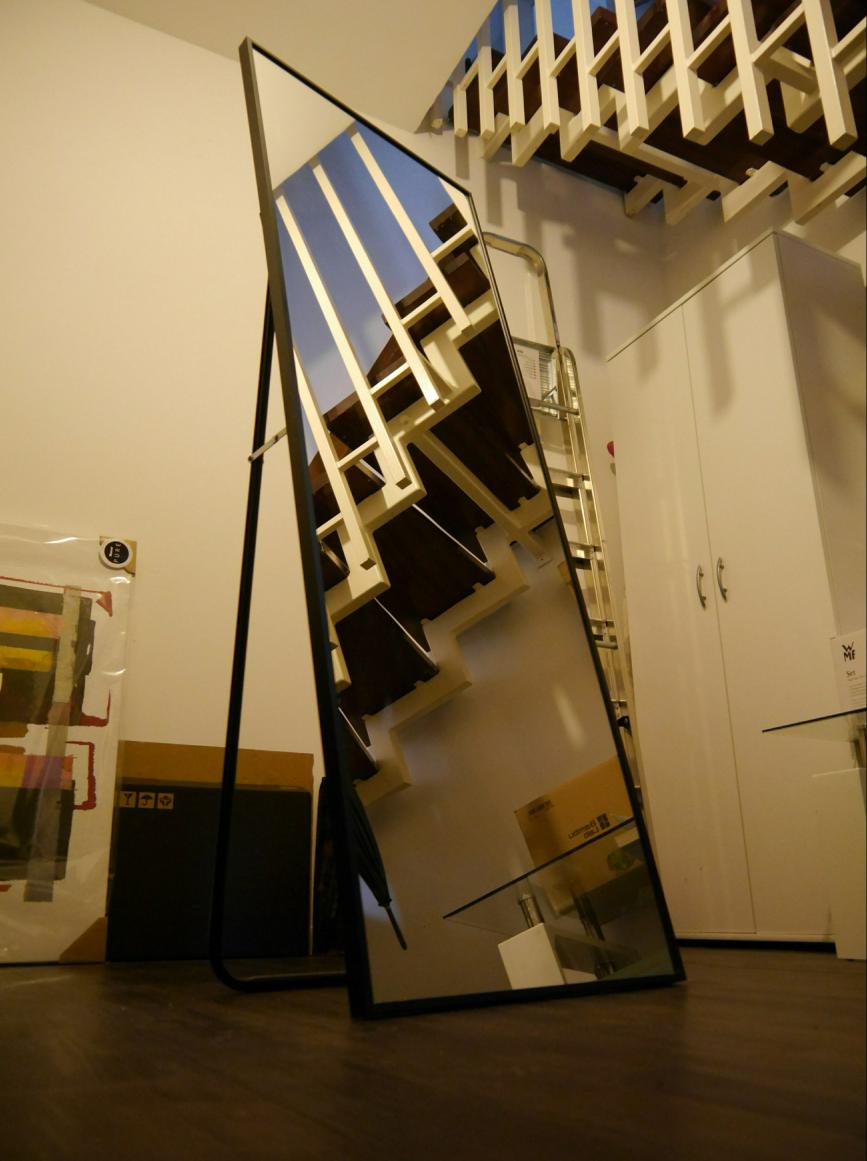}{0}{0.9}{-0.85}{-1.3}{0.8cm}{2}
    \end{subfigure}
    \\
    % %%%%%%%%%%%%
    \begin{subfigure}[t]{0.32\linewidth}
        \imagewithzoom{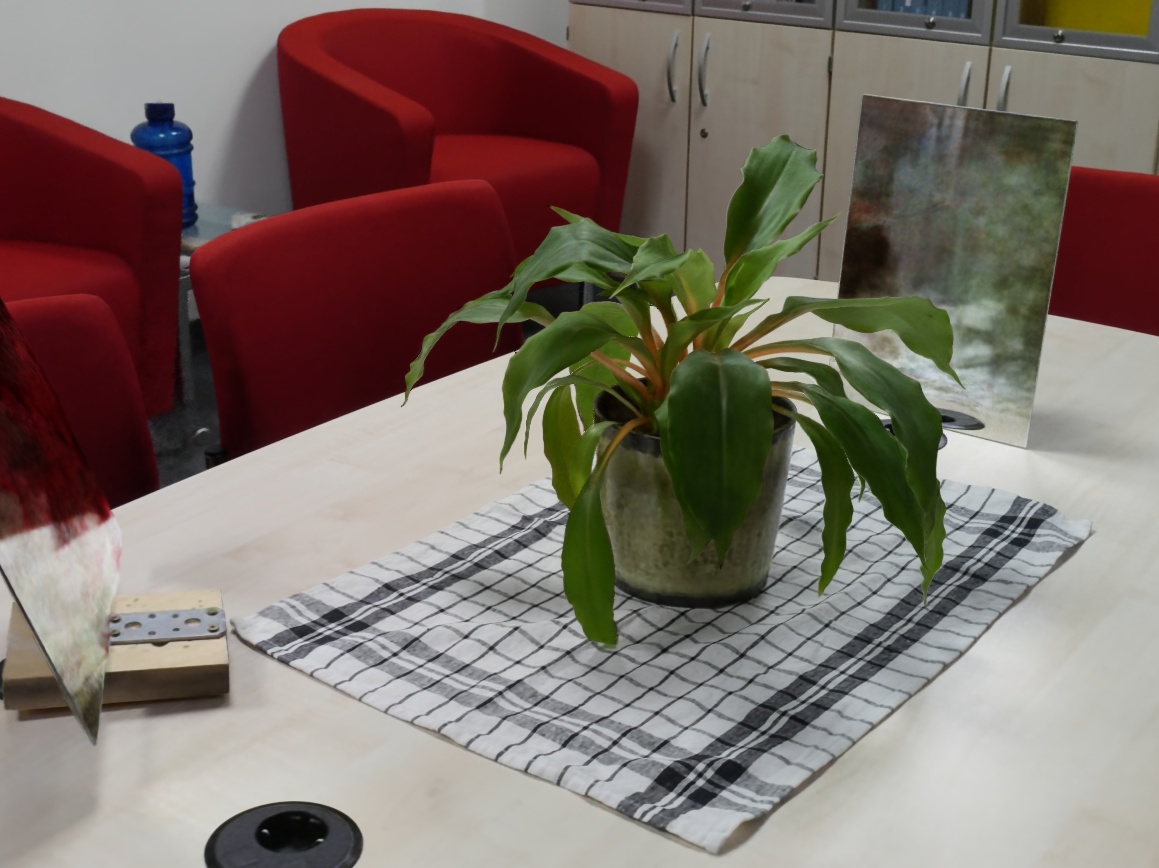}{1.0}{0.3}{-0.85}{0.55}{0.8cm}{3}
        \caption*{Best Baseline \\ Mip-NeRF 360 \cite{barron2022mipnerf360}}
    \end{subfigure}
    \hfill
    \begin{subfigure}[t]{0.32\linewidth}
        \imagewithzoom{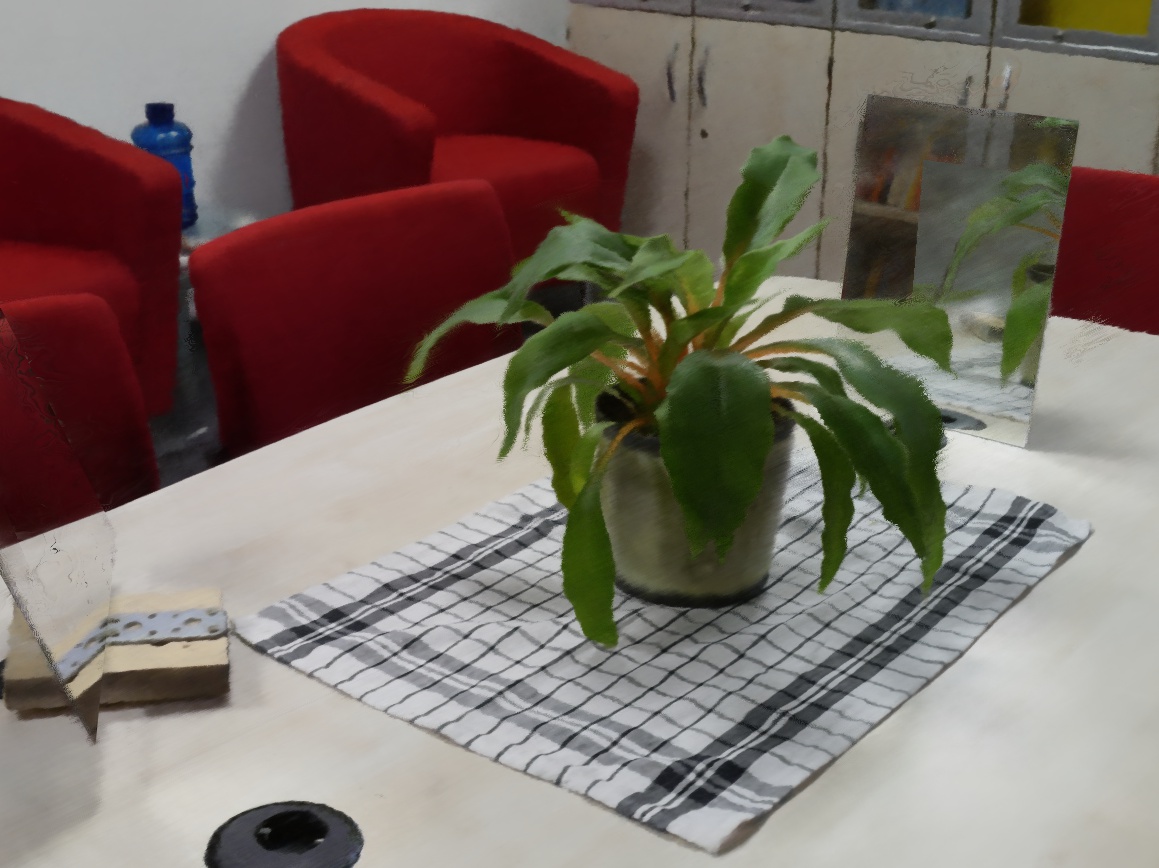}{1.0}{0.3}{-0.85}{0.55}{0.8cm}{3}
        \caption*{Ours}
    \end{subfigure}
    \hfill
    \begin{subfigure}[t]{0.32\linewidth}
        \imagewithzoom{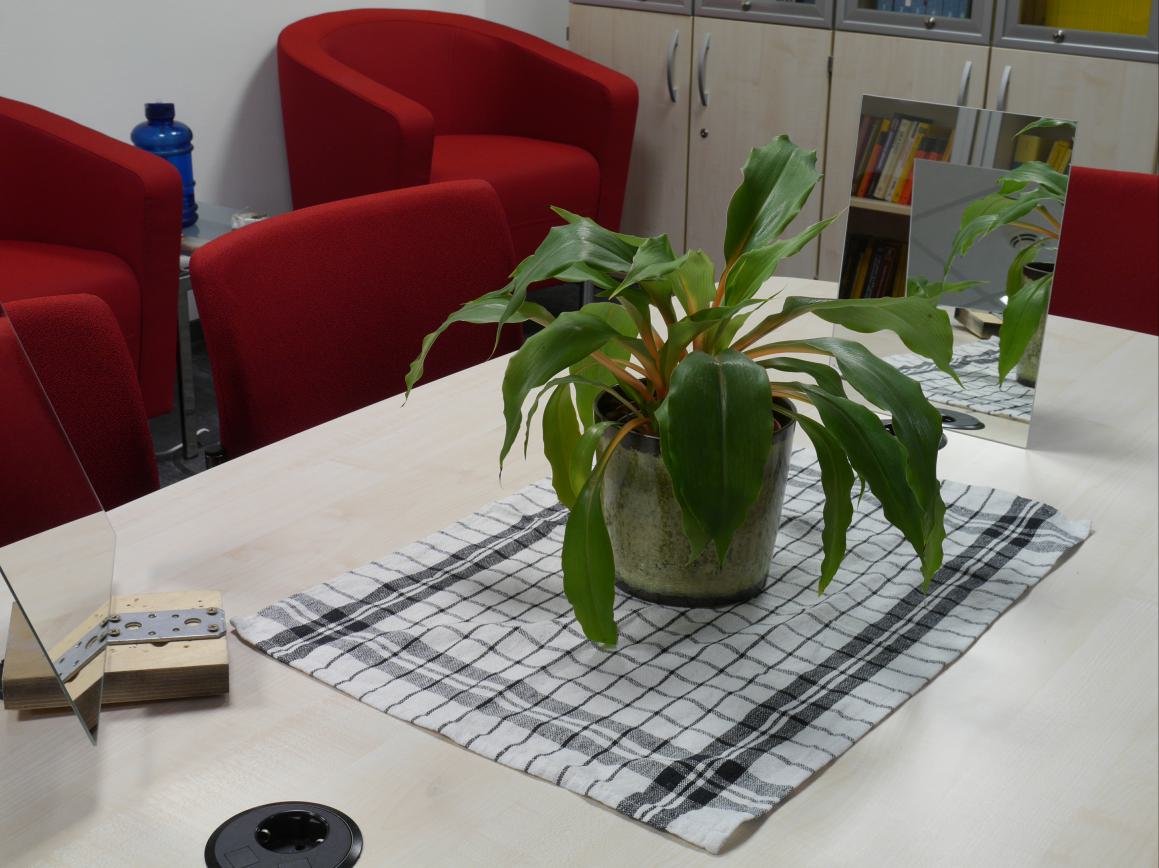}{1.0}{0.3}{-0.85}{0.55}{0.8cm}{3}
        \caption*{Ground Truth}
    \end{subfigure}
    \caption{Results on the test set of two real world scenes compared to the best baseline method according to \cref{tab:results}.}
    \label{fig:results_qualitative_real}
\end{figure}

\subsection{Synthetic Dataset}
The qualitative results on the synthetic scenes show a significant reduction of the blurriness artifacts the baseline methods inherit from the NeRF approach, as can be seen in \cref{fig:results_qualitative_synthetic}.
Techniques like MS-NeRF struggle to handle the scenario of high-order bounces, which becomes especially apparent in the bottom row of \cref{fig:results_qualitative_synthetic}, as the overall network capacity available is distributed into each of the learned subspaces, while our direct formulation can use the available information more efficiently if the total network capacity is equaled, as the colors observed in the reflections lead back to an optimization of the actual surface point.
This improvement can also be seen in the quantitative evaluation in \cref{tab:results} where our approach outperforms all baseline methods and even surpasses the results of TraM-NeRF on the LPIPS score of the full images.
Only if strong manual priors about the mirror location are provided, TraM-NeRF is able to yield significantly better results in the image metrics, but our method closely follows the quality and surpasses all other evaluated approaches.

\subsection{Real World Data}
While techniques like Mip-NeRF 360 can show their strengths in the overall reconstruction quality of real-world scenes, \cref{fig:results_qualitative_real} shows that it is prone to overfitting (top row; wall behind the mirror) and producing blurry artifacts in the presence of large mirror surfaces, whereas our approach detects and reconstructs the mirror surface in a plausible manner.
This strong baseline performance is also clear in the quantitative results in \cref{tab:results}, where Mip-NeRF 360 even outperforms the approaches using mirror priors.
However, TraM-NeRF outperforms the other approaches in the mirror regions, and we are able to closely follow its performance, however without relying on any annotations.

\section{Discussion}
While our method shows promising results for automatic detection of mirror surfaces in multi-view scenes, it also exhibits some limitations.
The detection depends on the lack of quality the standard NeRF approach produces in the presence of high-frequency multi-view inconsistencies.
In contrast, NeRF manages to reconstruct a mirror hanging on a wall if the area behind the wall is not captured, as the mirror image can just be reconstructed as proxy geometry in the empty space beyond the mirror surface.
We argue, however, that for sufficiently complex capture scenarios it becomes crucial to not rely on proxy geometry, especially if the geometric accuracy of the reconstruction is important for downstream tasks beyond novel view synthesis.
The current implementation of our approach is limited to the optimization of mirrors that can be seen by primary camera rays, as the antialiasing masks used for blending are only computed for the first surface interaction.
While it is easy to check for mirror intersections recursively if more than one mirror surface is present, reasoning about the area of the intersection of the reflected frustums is more involved, especially if part of the frustum interacted with a non-planar surface in the first bounce.
Future work is required to incorporate an approximation for the antialiasing for higher-order bounces in this general setting.
The shape fitting procedure described in \cref{sec:primitive_shape_fitting} relies on the apriori knowledge of the type and number of primitive shapes to detect.
While sophisticated shape detection methods for point clouds exist, this work focuses to derive quantities from standard NeRF reconstructions for mirror detection.
It would be interesting to further investigate such techniques \cite{schnabel2007ransac} in this context to reduce the number of involved hyperparameters.
\begin{figure}
    \centering
    \begin{subfigure}[t]{0.24\linewidth}
        \includegraphics[width=\linewidth]{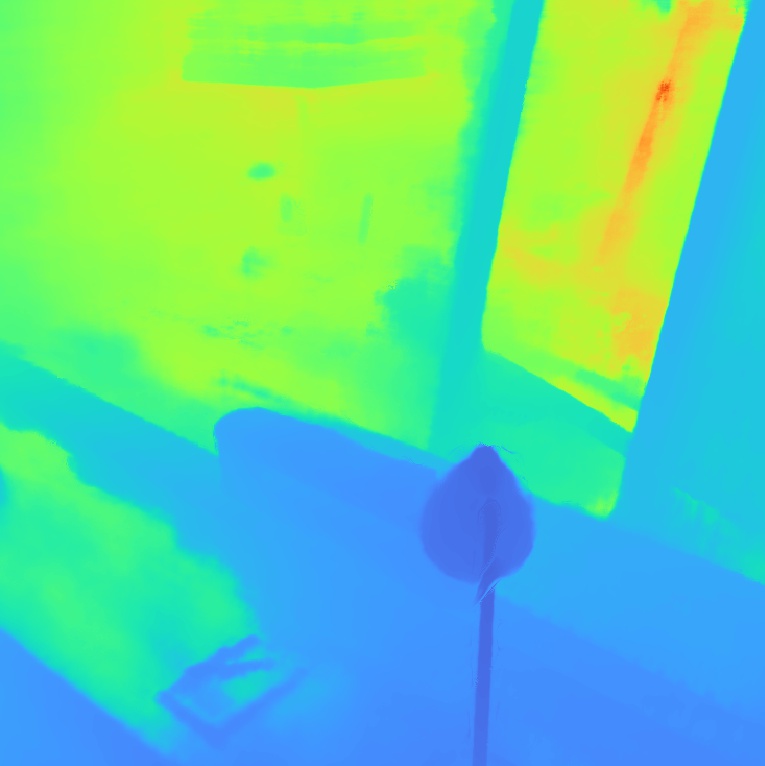}    
        \caption{$D(r)$}
    \end{subfigure}
    \hfill
    \begin{subfigure}[t]{0.24\linewidth}
        \includegraphics[width=\linewidth]{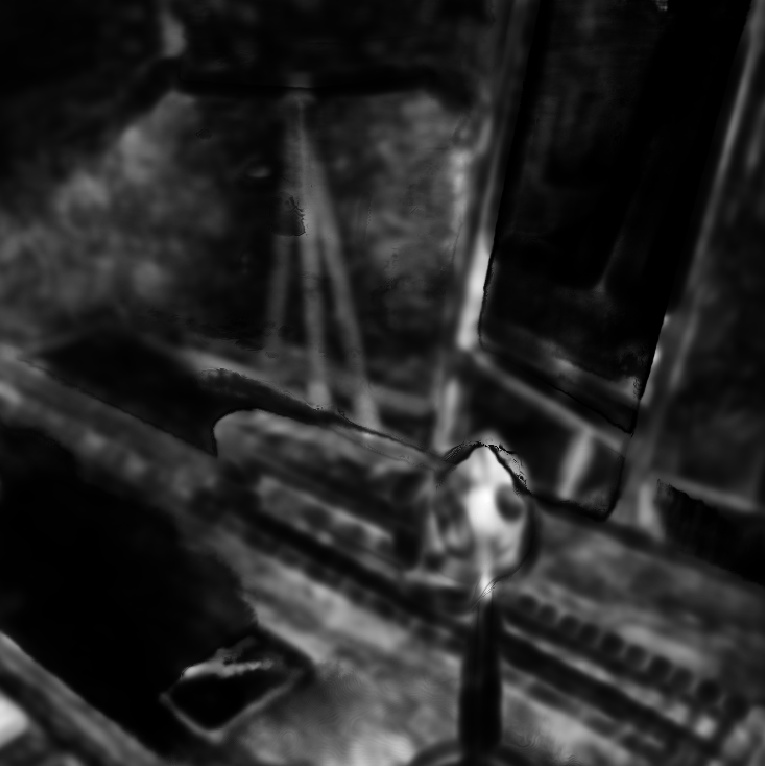}    
        \caption{$s(r)$}
    \end{subfigure}
    \hfill
    \begin{subfigure}[t]{0.24\linewidth}
        \includegraphics[width=\linewidth]{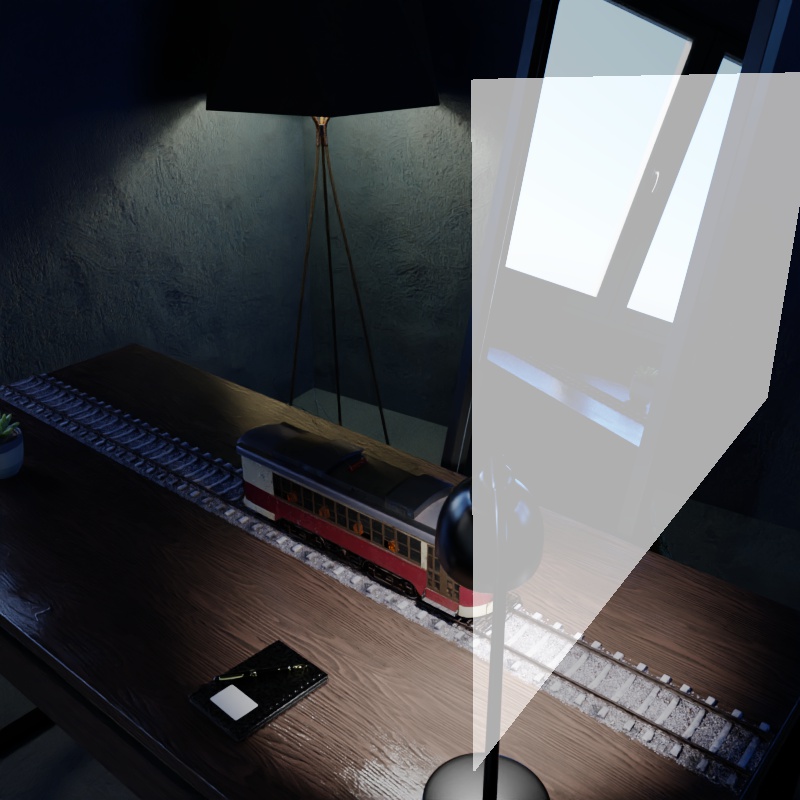}    
        \caption{Initial $\theta$}
    \end{subfigure}
    \hfill
    \begin{subfigure}[t]{0.24\linewidth}
        \imagewithzoom{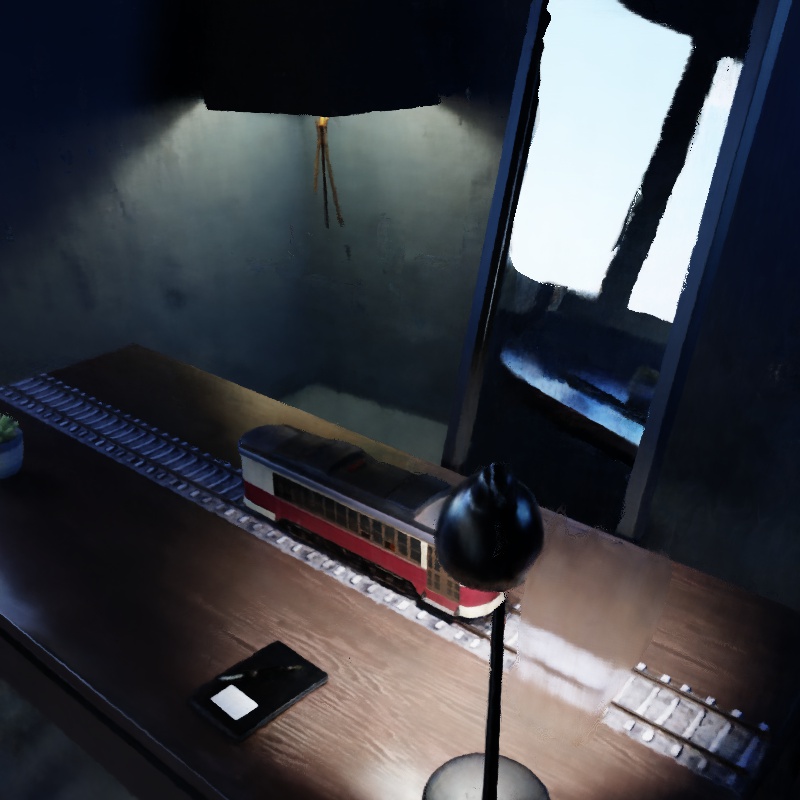}{0.42}{0.4}{-0.64}{0.64}{0.7cm}{1.8}
        \caption{Final RGB}
    \end{subfigure}

    \caption{Results on a scene from the TraM-NeRF dataset, showing the depth estimate (a), the weighted scores (b), the initial mirror parameters overlayed with the ground truth image (c) and the reconstruction result (d).}
    \label{fig:failure_case}
\end{figure}

\begin{figure}
\captionsetup[subfigure]{justification=centering}
\begin{subfigure}[t]{0.32\linewidth}
        \imagewithzoom{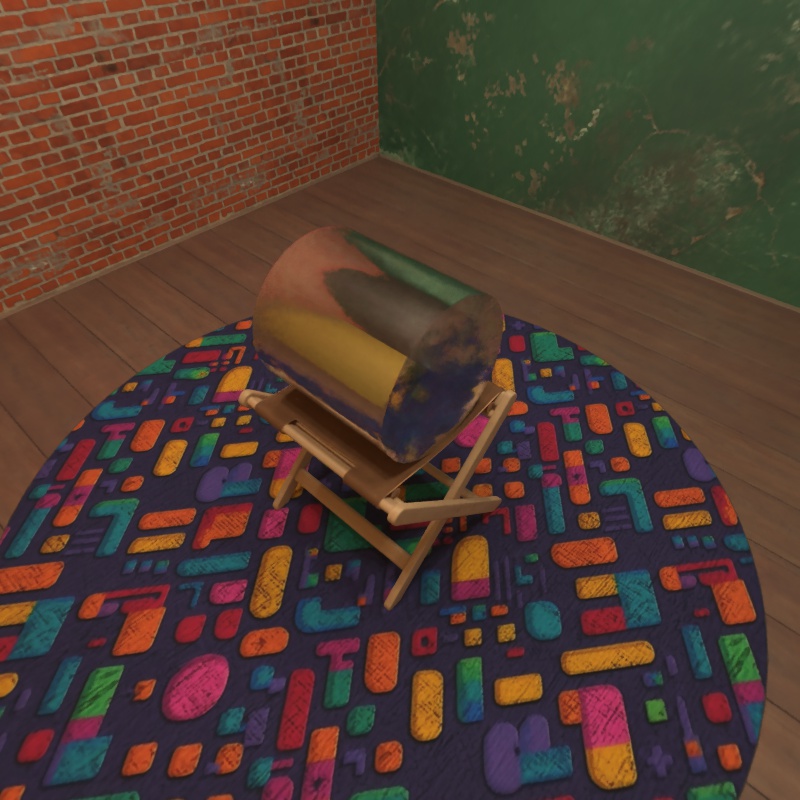}{0.1}{0.2}{-0.85}{0.85}{0.8cm}{3} % xleft yup
        \caption*{MS-NeRF \cite{yin2023msnerf} \\ PSNR: 33.55 dB}
    \end{subfigure}
    \hfill
    \begin{subfigure}[t]{0.32\linewidth}
    \imagewithzoom{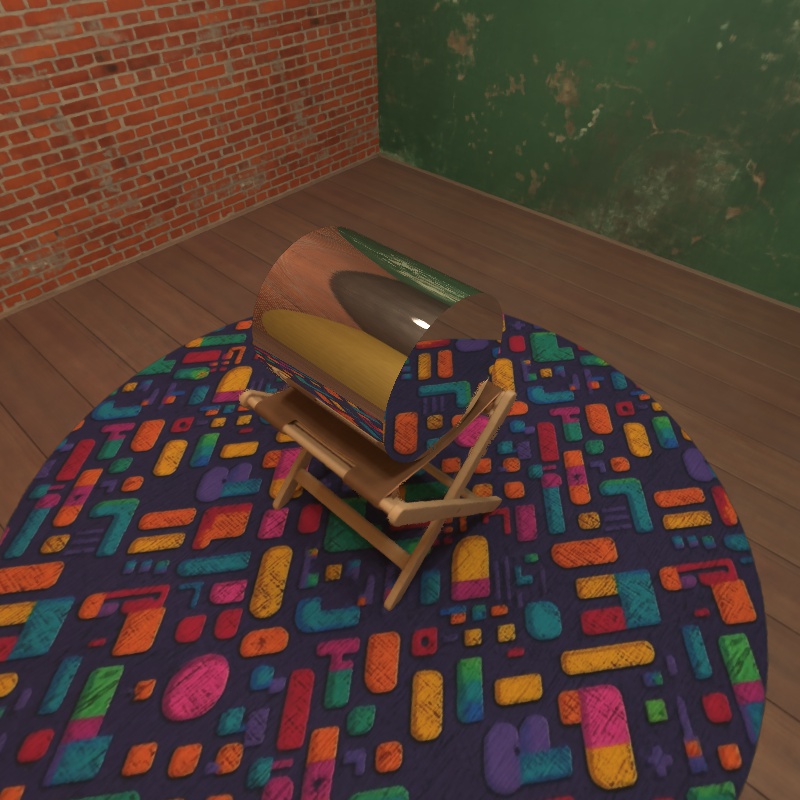}{0.1}{0.2}{-0.85}{0.85}{0.8cm}{3} % xleft yup
        \caption*{Ours \\ PSNR: 35.27 dB}
    \end{subfigure}
    \hfill
    \begin{subfigure}[t]{0.32\linewidth}
        \imagewithzoom{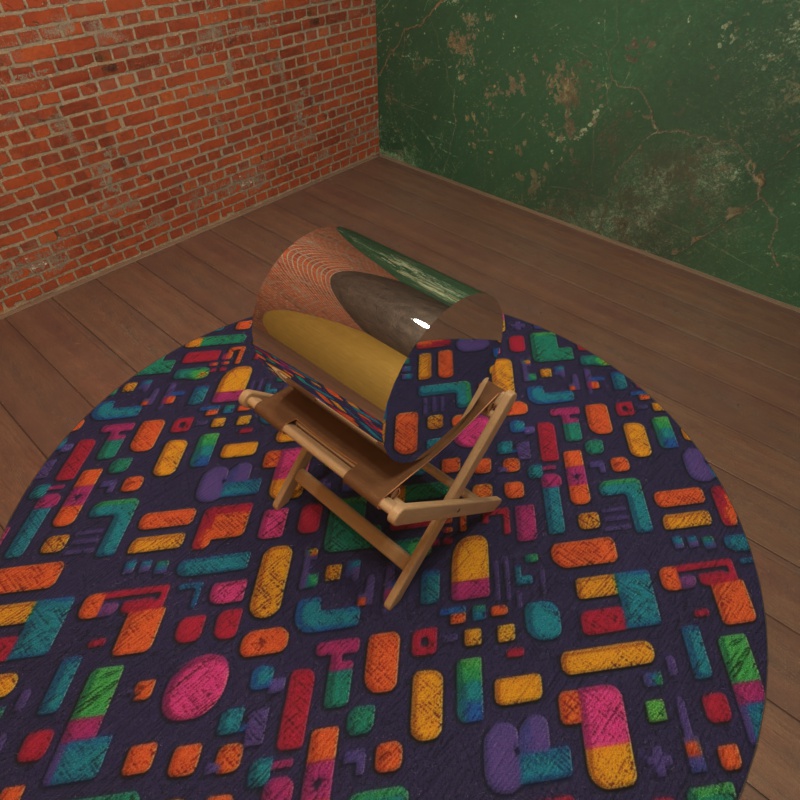}{0.1}{0.2}{-0.85}{0.85}{0.8cm}{3} % xleft yup
        \caption*{Ground Truth}
    \end{subfigure}
    \caption{Experimental results on the synthetic cylinder scene of the TraM-NeRF dataset \cite{holland2023tramnerf}.}
    \label{fig:results_cylinder}
\end{figure}
In case the shape detection fails and the mirror initialization is far away from the optimum, we observed that the optimization procedure quickly removes the influence of the erroneous reflections by shrinking the surface or moving it behind actual scene geometry.
That way, the results fall back to the quality of the underlying method, in this case achieving results close to Mip-NeRF.
An example of this is shown in \cref{fig:failure_case}, where the first step did not produce plausible depths for the mirror surface (a), leading to a mirror initialization far away from the actual location (c), such that the final result exhibits artifacts in the mirror region (d).
While we limited our experiments to scenes with planar mirrors, in \cref{fig:results_cylinder} we show that a synthetic cylinder scene~\cite{holland2023tramnerf} can also be successfully reconstructed by our approach.
However, further investigations are required to generalize the full method to other primitives, especially on real world data.

\section{Conclusion}
\label{sec:conclusion}

In this work, we presented NeRF-MD, an approach that analyses photometric inconsistencies present in the standard NeRF optimization to detect the presence of highly specular surfaces like mirrors and segments such regions to find and optimize an explicit surface representation using 3D primitives.
Our experiments substantiate that we are able to attain a significant improvement compared to baseline methods without the need of manual annotations of the mirror surfaces or ground truth masks.

\subsection*{Acknowledgements}
This work has been funded by the DFG project KL 1142/11-2 (DFG Research Unit FOR 2535 Anticipating Human Behaviour), and additionally by the Federal Ministry of Education and Research of Germany and the state of North-Rhine Westphalia as part of the Lamarr-Institute for Machine Learning and Artificial Intelligence and by the Federal Ministry of Education and Research under Grant No. 01IS22094E WEST-AI.

%%%%%%%%% REFERENCES
{\small
\bibliographystyle{ieee_fullname}
\bibliography{main}

\begin{thebibliography}{100}\itemsep=-1pt

\bibitem{attal2021torf}
Benjamin Attal, Eliot Laidlaw, Aaron Gokaslan, Changil Kim, Christian Richardt, James Tompkin, and Matthew O'Toole.
\newblock {TöRF}: Time-of-flight radiance fields for dynamic scene view synthesis.
\newblock {\em NeurIPS}, 34:26289--26301, 2021.

\bibitem{barron2021mipnerf}
Jonathan~T Barron, Ben Mildenhall, Matthew Tancik, Peter Hedman, Ricardo Martin-Brualla, and Pratul~P Srinivasan.
\newblock Mip-nerf: A multiscale representation for anti-aliasing neural radiance fields.
\newblock In {\em ICCV}, pages 5855--5864, 2021.

\bibitem{barron2022mipnerf360}
Jonathan~T Barron, Ben Mildenhall, Dor Verbin, Pratul~P Srinivasan, and Peter Hedman.
\newblock Mip-nerf 360: Unbounded anti-aliased neural radiance fields.
\newblock In {\em CVPR}, pages 5470--5479, 2022.

\bibitem{barron2023zip}
Jonathan~T Barron, Ben Mildenhall, Dor Verbin, Pratul~P Srinivasan, and Peter Hedman.
\newblock Zip-nerf: Anti-aliased grid-based neural radiance fields.
\newblock In {\em ICCV}, 2023.

\bibitem{bi2020neural}
Sai Bi, Zexiang Xu, Pratul Srinivasan, Ben Mildenhall, Kalyan Sunkavalli, Milo{\v{s}} Ha{\v{s}}an, Yannick Hold-Geoffroy, David Kriegman, and Ravi Ramamoorthi.
\newblock Neural reflectance fields for appearance acquisition.
\newblock {\em arXiv preprint arXiv:2008.03824}, 2020.

\bibitem{bi2020deep}
Sai Bi, Zexiang Xu, Kalyan Sunkavalli, Milo{\v{s}} Ha{\v{s}}an, Yannick Hold-Geoffroy, David Kriegman, and Ravi Ramamoorthi.
\newblock Deep reflectance volumes: Relightable reconstructions from multi-view photometric images.
\newblock In {\em ECCV}, pages 294--311. Springer, 2020.

\bibitem{Bian-2022-NoPe-NeRF}
Wenjing Bian, Zirui Wang, Kejie Li, Jia-Wang Bian, and Victor~Adrian Prisacariu.
\newblock {NoPe-NeRF}: Optimising neural radiance field with no pose prior.
\newblock In {\em CVPR}, pages 4160--4169, 2023.

\bibitem{boss2021nerd}
Mark Boss, Raphael Braun, Varun Jampani, Jonathan~T. Barron, Ce Liu, and Hendrik Lensch.
\newblock Nerd: Neural reflectance decomposition from image collections.
\newblock In {\em ICCV}, pages 12684--12694, 2021.

\bibitem{boss2022samurai}
Mark Boss, Andreas Engelhardt, Abhishek Kar, Yuanzhen Li, Deqing Sun, Jonathan Barron, Hendrik Lensch, and Varun Jampani.
\newblock Samurai: Shape and material from unconstrained real-world arbitrary image collections.
\newblock {\em NeurIPS}, 35:26389--26403, 2022.

\bibitem{boss2021neural}
Mark Boss, Varun Jampani, Raphael Braun, Ce Liu, Jonathan Barron, and Hendrik Lensch.
\newblock Neural-pil: Neural pre-integrated lighting for reflectance decomposition.
\newblock {\em NeurIPS}, 34:10691--10704, 2021.

\bibitem{Cao2022rtNeRF}
Junli Cao, Huan Wang, Pavlo Chemerys, Vladislav Shakhrai, Ju Hu, Yun Fu, Denys Makoviichuk, Sergey Tulyakov, and Jian Ren.
\newblock Real-time neural light field on mobile devices.
\newblock In {\em CVPR}, pages 8328--8337, 2023.

\bibitem{chen2022tensorf}
Anpei Chen, Zexiang Xu, Andreas Geiger, Jingyi Yu, and Hao Su.
\newblock Tensorf: Tensorial radiance fields.
\newblock In {\em ECCV}, pages 333--350, 2022.

\bibitem{chen2021animatable}
Jianchuan Chen, Ying Zhang, Di Kang, Xuefei Zhe, Linchao Bao, Xu Jia, and Huchuan Lu.
\newblock Animatable neural radiance fields from monocular rgb videos.
\newblock {\em arXiv preprint arXiv:2106.13629}, 6 2021.

\bibitem{chen2022hallucinated}
Xingyu Chen, Qi Zhang, Xiaoyu Li, Yue Chen, Ying Feng, Xuan Wang, and Jue Wang.
\newblock Hallucinated neural radiance fields in the wild.
\newblock In {\em CVPR}, pages 12943--12952, 2022.

\bibitem{Chen-2023-L2G-NeRF}
Yue Chen, Xingyu Chen, Xuan Wang, Qi Zhang, Yu Guo, Ying Shan, and Fei Wang.
\newblock Local-to-global registration for bundle-adjusting neural radiance fields.
\newblock In {\em CVPR}, pages 8264--8273, 2023.

\bibitem{chen-2023-DBARF}
Yu Chen and Gim~Hee Lee.
\newblock {DBARF:} deep bundle-adjusting generalizable neural radiance fields.
\newblock In {\em CVPR}, pages 24--34, 2023.

\bibitem{chen2022mobilenerf}
Zhiqin Chen, Thomas Funkhouser, Peter Hedman, and Andrea Tagliasacchi.
\newblock {MobileNeRF}: Exploiting the polygon rasterization pipeline for efficient neural field rendering on mobile architectures.
\newblock In {\em CVPR}, pages 16569--16578, 2023.

\bibitem{Cheng20223LUNeRF}
Zezhou Cheng, Carlos Esteves, Varun Jampani, Abhishek Kar, Subhransu Maji, and Ameesh Makadia.
\newblock {LU-NeRF}: Scene and pose estimation by synchronizing local unposed nerfs.
\newblock In {\em ICCV}, 2023.

\bibitem{cho2022streamable}
Junwoo Cho, Seungtae Nam, Daniel Rho, Jong~Hwan Ko, and Eunbyung Park.
\newblock Streamable neural fields.
\newblock In Shai Avidan, Gabriel~J. Brostow, Moustapha Cissé, Giovanni~Maria Farinella, and Tal Hassner, editors, {\em ECCV}, volume 13680, pages 595--612. Springer, Springer Nature Switzerland, 7 2022.

\bibitem{Chung-2022-Orbeez-SLAM}
Chi-Ming Chung, Yang-Che Tseng, Ya-Ching Hsu, Xiang-Qian Shi, Yun-Hung Hua, Jia-Fong Yeh, Wen-Chin Chen, Yi-Ting Chen, and Winston~H Hsu.
\newblock {Orbeez-SLAM}: {A} real-time monocular visual {SLAM} with {ORB} features and nerf-realized mapping.
\newblock In {\em ICRA}, pages 9400--9406. IEEE, 2023.

\bibitem{deng2022depth}
Kangle Deng, Andrew Liu, Jun-Yan Zhu, and Deva Ramanan.
\newblock Depth-supervised {NeRF}: Fewer views and faster training for free.
\newblock In {\em CVPR}, pages 12882--12891. IEEE, 6 2022.

\bibitem{du2021neural}
Yilun Du, Yinan Zhang, Hong-Xing Yu, Joshua~B Tenenbaum, and Jiajun Wu.
\newblock Neural radiance flow for {4D} view synthesis and video processing.
\newblock In {\em ICCV}, pages 14304--14314. IEEE Computer Society, IEEE, 10 2021.

\bibitem{fan2023factored}
Yue Fan, Ivan Skorokhodov, Oleg Voynov, Savva Ignatyev, Evgeny Burnaev, Peter Wonka, and Yiqun Wang.
\newblock Factored-neus: Reconstructing surfaces, illumination, and materials of possibly glossy objects.
\newblock {\em arXiv preprint arXiv:2305.17929}, 2023.

\bibitem{fang2022fast}
Jiemin Fang, Taoran Yi, Xinggang Wang, Lingxi Xie, Xiaopeng Zhang, Wenyu Liu, Matthias Nie{\ss}ner, and Qi Tian.
\newblock Fast dynamic radiance fields with time-aware neural voxels.
\newblock In {\em SIGGRAPH Asia}, pages 1--9, 2022.

\bibitem{fischler1981random}
Martin~A Fischler and Robert~C Bolles.
\newblock Random sample consensus: a paradigm for model fitting with applications to image analysis and automated cartography.
\newblock {\em Communications of the ACM}, 24(6):381--395, 1981.

\bibitem{fridovich2022plenoxels}
Sara Fridovich-Keil, Alex Yu, Matthew Tancik, Qinhong Chen, Benjamin Recht, and Angjoo Kanazawa.
\newblock Plenoxels: Radiance fields without neural networks.
\newblock In {\em CVPR}, pages 5501--5510, 2022.

\bibitem{gafni2021dynamic}
Guy Gafni, Justus Thies, Michael Zollhofer, and Matthias Nießner.
\newblock Dynamic neural radiance fields for monocular {4D} facial avatar reconstruction.
\newblock In {\em CVPR}, pages 8649--8658. IEEE, 6 2021.

\bibitem{gao2021dynamic}
Chen Gao, Ayush Saraf, Johannes Kopf, and Jia-Bin Huang.
\newblock Dynamic view synthesis from dynamic monocular video.
\newblock In {\em ICCV}, pages 5712--5721. IEEE, 10 2021.

\bibitem{garbin2021fastnerf}
Stephan~J Garbin, Marek Kowalski, Matthew Johnson, Jamie Shotton, and Julien Valentin.
\newblock Fastnerf: High-fidelity neural rendering at 200fps.
\newblock In {\em ICCV}, pages 14346--14355, 2021.

\bibitem{ge2023ref}
Wenhang Ge, Tao Hu, Haoyu Zhao, Shu Liu, and Ying-Cong Chen.
\newblock Ref-neus: Ambiguity-reduced neural implicit surface learning for multi-view reconstruction with reflection.
\newblock In {\em ICCV}, 2023.

\bibitem{guo2022nerfren}
Yuan-Chen Guo, Di Kang, Linchao Bao, Yu He, and Song-Hai Zhang.
\newblock Nerfren: Neural radiance fields with reflections.
\newblock In {\em CVPR}, pages 18409--18418, 2022.

\bibitem{Heo-2023-CamPos_MResHashEncoding}
Hwan Heo, Taekyung Kim, Jiyoung Lee, Jaewon Lee, Soohyun Kim, Hyunwoo~J. Kim, and Jin{-}Hwa Kim.
\newblock Robust camera pose refinement for multi-resolution hash encoding.
\newblock In {\em ICML}, 2023.

\bibitem{holland2023tramnerf}
Leif~Van Holland, Ruben Bliersbach, Jan~Uwe M{\" u}ller, Patrick Stotko, and Reinhard Klein.
\newblock Tram-{NeRF}: Tracing {Mirror} and {Near}-{Perfect} {Specular} {Reflections} {Through} {Neural} {Radiance} {Fields}.
\newblock {\em Computer Graphics Forum}, 2024.

\bibitem{huang2022hdr}
Xin Huang, Qi Zhang, Ying Feng, Hongdong Li, Xuan Wang, and Qing Wang.
\newblock Hdr-nerf: High dynamic range neural radiance fields.
\newblock In {\em CVPR}, pages 18398--18408, 2022.

\bibitem{jeong2021self}
Yoonwoo Jeong, Seokjun Ahn, Christopher Choy, Anima Anandkumar, Minsu Cho, and Jaesik Park.
\newblock Self-calibrating neural radiance fields.
\newblock In {\em ICCV}, pages 5846--5854. IEEE, 10 2021.

\bibitem{Jiang2022alignnerf}
Yifan Jiang, Peter Hedman, Ben Mildenhall, Dejia Xu, Jonathan~T Barron, Zhangyang Wang, and Tianfan Xue.
\newblock {AligNeRF}: High-fidelity neural radiance fields via alignment-aware training.
\newblock In {\em CVPR}, pages 46--55, 2023.

\bibitem{jin2023tensoir}
Haian Jin, Isabella Liu, Peijia Xu, Xiaoshuai Zhang, Songfang Han, Sai Bi, Xiaowei Zhou, Zexiang Xu, and Hao Su.
\newblock Tensoir: Tensorial inverse rendering.
\newblock In {\em CVPR}, pages 165--174, 2023.

\bibitem{Seong2022hdrplenoxels}
Kim Jun-Seong, Kim Yu-Ji, Moon Ye-Bin, and Tae-Hyun Oh.
\newblock {HDR-P}lenoxels: Self-calibrating high dynamic range radiance fields.
\newblock In {\em ECCV}, pages 384--401. Springer, 2022.

\bibitem{kerbl20233d}
Bernhard Kerbl, Georgios Kopanas, Thomas Leimk{\"u}hler, and George Drettakis.
\newblock 3d gaussian splatting for real-time radiance field rendering.
\newblock {\em TOG}, 42(4):1--14, 2023.

\bibitem{kuang2022neroic}
Zhengfei Kuang, Kyle Olszewski, Menglei Chai, Zeng Huang, Panos Achlioptas, and Sergey Tulyakov.
\newblock Neroic: Neural rendering of objects from online image collections.
\newblock {\em TOG}, 41(4):1--12, 2022.

\bibitem{Laine2020diffrast}
Samuli Laine, Janne Hellsten, Tero Karras, Yeongho Seol, Jaakko Lehtinen, and Timo Aila.
\newblock Modular primitives for high-performance differentiable rendering.
\newblock {\em ACM Transactions on Graphics}, 39(6), 2020.

\bibitem{Li2022streaming}
Lingzhi Li, Zhen Shen, Zhongshu Wang, Li Shen, and Ping Tan.
\newblock Streaming radiance fields for 3d video synthesis.
\newblock In {\em NeurIPS}, 2022.

\bibitem{li2023uhdnerf}
Quewei Li, Feichao Li, Jie Guo, and Yanwen Guo.
\newblock Uhdnerf: Ultra-high-definition neural radiance fields.
\newblock In {\em ICCV}, pages 23097--23108, 2023.

\bibitem{li2022neural}
Tianye Li, Mira Slavcheva, Michael Zollhoefer, et~al.
\newblock Neural {3D} video synthesis from multi-view video.
\newblock In {\em CVPR}, pages 5521--5531. IEEE, 6 2022.

\bibitem{li2021neural}
Zhengqi Li, Simon Niklaus, Noah Snavely, and Oliver Wang.
\newblock Neural scene flow fields for space-time view synthesis of dynamic scenes.
\newblock In {\em CVPR}, pages 6498--6508. IEEE, 6 2021.

\bibitem{Li2022DynIBaR}
Zhengqi Li, Qianqian Wang, Forrester Cole, Richard Tucker, and Noah Snavely.
\newblock {DynIBaR:} neural dynamic image-based rendering.
\newblock In {\em CVPR}, pages 4273--4284, 2023.

\bibitem{liang2023envidr}
Ruofan Liang, Huiting Chen, Chunlin Li, Fan Chen, Selvakumar Panneer, and Nandita Vijaykumar.
\newblock Envidr: Implicit differentiable renderer with neural environment lighting.
\newblock In {\em ICCV}, 2023.

\bibitem{lin2021barf}
Chen-Hsuan Lin, Wei-Chiu Ma, Antonio Torralba, and Simon Lucey.
\newblock {BaRF}: Bundle-adjusting neural radiance fields.
\newblock In {\em ICCV}, pages 5741--5751. IEEE, 10 2021.

\bibitem{Liu2023NeRF-Loc}
Jianlin Liu, Qiang Nie, Yong Liu, and Chengjie Wang.
\newblock {NeRF-Loc}: Visual localization with conditional neural radiance field.
\newblock In {\em ICRA}, 2023.

\bibitem{liu2024mirrorgaussian}
Jiayue Liu, Xiao Tang, Freeman Cheng, Roy Yang, Zhihao Li, Jianzhuang Liu, Yi Huang, Jiaqi Lin, Shiyong Liu, Xiaofei Wu, et~al.
\newblock Mirrorgaussian: Reflecting 3d gaussians for reconstructing mirror reflections.
\newblock {\em arXiv preprint arXiv:2405.11921}, 2024.

\bibitem{liu2021neural}
Lingjie Liu, Marc Habermann, Viktor Rudnev, Kripasindhu Sarkar, Jiatao Gu, and Christian Theobalt.
\newblock Neural actor: Neural free-view synthesis of human actors with pose control.
\newblock {\em TOG}, 40(6):1--16, 2021.

\bibitem{lloyd1982least}
Stuart Lloyd.
\newblock Least squares quantization in pcm.
\newblock {\em IEEE transactions on information theory}, 28(2):129--137, 1982.

\bibitem{lombardi2019neural}
Stephen Lombardi, Tomas Simon, Jason Saragih, Gabriel Schwartz, Andreas Lehrmann, and Yaser Sheikh.
\newblock Neural volumes: learning dynamic renderable volumes from images.
\newblock {\em TOG}, 38(4):1--14, 2019.

\bibitem{ma2023specnerf}
Li Ma, Vasu Agrawal, Haithem Turki, Changil Kim, Chen Gao, Pedro Sander, Michael Zollh{\"o}fer, and Christian Richardt.
\newblock Specnerf: Gaussian directional encoding for specular reflections.
\newblock {\em arXiv preprint arXiv:2312.13102}, 2023.

\bibitem{Maggio-2023-Loc-NeRF}
Dominic Maggio, Marcus Abate, Jingnan Shi, Courtney Mario, and Luca Carlone.
\newblock {Loc-NeRF}: Monte carlo localization using neural radiance fields.
\newblock {\em ICRA}, pages 4018--4025, 2023.

\bibitem{mai2023neural}
Alexander Mai, Dor Verbin, Falko Kuester, and Sara Fridovich-Keil.
\newblock Neural microfacet fields for inverse rendering.
\newblock In {\em ICCV}, pages 408--418, 2023.

\bibitem{martin2021nerf}
Ricardo Martin-Brualla, Noha Radwan, Mehdi~SM Sajjadi, Jonathan~T Barron, Alexey Dosovitskiy, and Daniel Duckworth.
\newblock Nerf in the wild: Neural radiance fields for unconstrained photo collections.
\newblock In {\em CVPR}, pages 7210--7219, 2021.

\bibitem{max1995optical}
Nelson Max.
\newblock Optical models for direct volume rendering.
\newblock {\em TVCG}, 1(2):99--108, 1995.

\bibitem{meng2024mirror}
Jiarui Meng, Haijie Li, Yanmin Wu, Qiankun Gao, Shuzhou Yang, Jian Zhang, and Siwei Ma.
\newblock Mirror-3dgs: Incorporating mirror reflections into 3d gaussian splatting.
\newblock {\em arXiv preprint arXiv:2404.01168}, 2024.

\bibitem{meng2021gnerf}
Quan Meng, Anpei Chen, Haimin Luo, Minye Wu, Hao Su, Lan Xu, Xuming He, and Jingyi Yu.
\newblock {GNeRF}: {GAN}-based neural radiance field without posed camera.
\newblock In {\em ICCV}, pages 6351--6361. IEEE, 10 2021.

\bibitem{Mi2023switchnerf}
Zhenxing Mi and Dan Xu.
\newblock {Switch-NeRF}: Learning scene decomposition with mixture of experts for large-scale neural radiance fields.
\newblock In {\em ICLR}, 2023.

\bibitem{mildenhall2022nerf}
Ben Mildenhall, Peter Hedman, Ricardo Martin-Brualla, Pratul~P Srinivasan, and Jonathan~T Barron.
\newblock Nerf in the dark: High dynamic range view synthesis from noisy raw images.
\newblock In {\em CVPR}, pages 16190--16199, 2022.

\bibitem{mildenhall2020nerf}
Ben Mildenhall, Pratul~P. Srinivasan, Matthew Tancik, Jonathan~T. Barron, Ravi Ramamoorthi, and Ren Ng.
\newblock Nerf: Representing scenes as neural radiance fields for view synthesis.
\newblock In {\em ECCV}, 2020.

\bibitem{Mubarik2023HardwareAcc}
Muhammad~Husnain Mubarik, Ramakrishna Kanungo, Tobias Zirr, and Rakesh Kumar.
\newblock Hardware acceleration of neural graphics.
\newblock In Yan Solihin and Mark~A. Heinrich, editors, {\em ISCA}, pages 50:1--50:12. {ACM}, 2023.

\bibitem{mueller2022instant}
Thomas M\"uller, Alex Evans, Christoph Schied, and Alexander Keller.
\newblock Instant neural graphics primitives with a multiresolution hash encoding.
\newblock {\em TOG}, 41(4):102:1--102:15, 2022.

\bibitem{munkberg2022extracting}
Jacob Munkberg, Jon Hasselgren, Tianchang Shen, Jun Gao, Wenzheng Chen, Alex Evans, Thomas M{\"u}ller, and Sanja Fidler.
\newblock Extracting triangular 3d models, materials, and lighting from images.
\newblock In {\em CVPR}, pages 8280--8290, 2022.

\bibitem{niemeyer2021giraffe}
Michael Niemeyer and Andreas Geiger.
\newblock Giraffe: Representing scenes as compositional generative neural feature fields.
\newblock In {\em CVPR}, pages 11453--11464, 2021.

\bibitem{niemeyer2020differentiable}
Michael Niemeyer, Lars Mescheder, Michael Oechsle, and Andreas Geiger.
\newblock Differentiable volumetric rendering: Learning implicit 3d representations without 3d supervision.
\newblock In {\em CVPR}, pages 3504--3515, 2020.

\bibitem{noguchi2021neural}
Atsuhiro Noguchi, Xiao Sun, Stephen Lin, and Tatsuya Harada.
\newblock Neural articulated radiance field.
\newblock In {\em ICCV}, pages 5762--5772. IEEE, 10 2021.

\bibitem{park2021nerfies}
Keunhong Park, Utkarsh Sinha, Jonathan~T Barron, Sofien Bouaziz, Dan~B Goldman, Steven~M Seitz, and Ricardo Martin-Brualla.
\newblock Nerfies: Deformable neural radiance fields.
\newblock In {\em ICCV}, pages 5865--5874. IEEE, 10 2021.

\bibitem{park2021hypernerf}
Keunhong Park, Utkarsh Sinha, Peter Hedman, Jonathan~T Barron, Sofien Bouaziz, Dan~B Goldman, Ricardo Martin-Brualla, and Steven~M Seitz.
\newblock {HyperNeRF}: A higher-dimensional representation for topologically varying neural radiance fields.
\newblock {\em TOG}, 2021.

\bibitem{peng2021animatable}
Sida Peng, Junting Dong, Qianqian Wang, Shangzhan Zhang, Qing Shuai, Xiaowei Zhou, and Hujun Bao.
\newblock Animatable neural radiance fields for modeling dynamic human bodies.
\newblock In {\em ICCV}, pages 14314--14323. IEEE, 10 2021.

\bibitem{peng2021neural}
Sida Peng, Yuanqing Zhang, Yinghao Xu, Qianqian Wang, Qing Shuai, Hujun Bao, and Xiaowei Zhou.
\newblock Neural body: Implicit neural representations with structured latent codes for novel view synthesis of dynamic humans.
\newblock In {\em CVPR}, pages 9054--9063. IEEE, 6 2021.

\bibitem{pumarola2021d}
Albert Pumarola, Enric Corona, Gerard Pons-Moll, and Francesc Moreno-Noguer.
\newblock {D-NeRF}: Neural radiance fields for dynamic scenes.
\newblock In {\em CVPR}, pages 10318--10327. IEEE, 6 2021.

\bibitem{raj2021}
Amit Raj, Michael Zollhöfer, Tomas Simon, Jason~M. Saragih, Shunsuke Saito, James Hays, and Stephen Lombardi.
\newblock Pixel-aligned volumetric avatars.
\newblock In {\em CVPR}, pages 11733--11742. IEEE, 6 2021.

\bibitem{reiser2021kilonerf}
Christian Reiser, Songyou Peng, Yiyi Liao, and Andreas Geiger.
\newblock Kilonerf: Speeding up neural radiance fields with thousands of tiny mlps.
\newblock In {\em ICCV}, pages 14335--14345, 2021.

\bibitem{rematas2022urban}
Konstantinos Rematas, Andrew Liu, Pratul~P Srinivasan, Jonathan~T Barron, Andrea Tagliasacchi, Thomas Funkhouser, and Vittorio Ferrari.
\newblock Urban radiance fields.
\newblock In {\em CVPR}, pages 12932--12942. IEEE, 6 2022.

\bibitem{roessle2022dense}
Barbara Roessle, Jonathan~T Barron, Ben Mildenhall, Pratul~P Srinivasan, and Matthias Nießner.
\newblock Dense depth priors for neural radiance fields from sparse input views.
\newblock In {\em CVPR}, pages 12892--12901. IEEE, 6 2022.

\bibitem{rosinol2022nerf}
Antoni Rosinol, John~J Leonard, and Luca Carlone.
\newblock {NeRF-SLAM}: Real-time dense monocular {SLAM} with neural radiance fields.
\newblock {\em arXiv preprint arXiv:2210.13641}, 2022.

\bibitem{schnabel2007ransac}
R. Schnabel, R. Wahl, and R. Klein.
\newblock Efficient ransac for point-cloud shape detection.
\newblock {\em Computer Graphics Forum}, 26(2):214--226, 2007.

\bibitem{sitzmann2019scene}
Vincent Sitzmann, Michael Zollh{\"o}fer, and Gordon Wetzstein.
\newblock Scene representation networks: Continuous 3d-structure-aware neural scene representations.
\newblock {\em NeurIPS}, 32, 2019.

\bibitem{srinivasan2021nerv}
Pratul~P Srinivasan, Boyang Deng, Xiuming Zhang, Matthew Tancik, Ben Mildenhall, and Jonathan~T Barron.
\newblock Nerv: Neural reflectance and visibility fields for relighting and view synthesis.
\newblock In {\em CVPR}, pages 7495--7504, 2021.

\bibitem{sucar2021imap}
Edgar Sucar, Shikun Liu, Joseph Ortiz, and Andrew~J Davison.
\newblock {iMAP}: Implicit mapping and positioning in real-time.
\newblock In {\em ICCV}, pages 6229--6238. IEEE, 10 2021.

\bibitem{sun2022direct}
Cheng Sun, Min Sun, and Hwann-Tzong Chen.
\newblock Direct voxel grid optimization: Super-fast convergence for radiance fields reconstruction.
\newblock In {\em CVPR}, pages 5459--5469. IEEE, 6 2022.

\bibitem{Sun2023NeRF-Loc}
Jiankai Sun, Yan Xu, Mingyu Ding, Hongwei Yi, Chen Wang, Jingdong Wang, Liangjun Zhang, and Mac Schwager.
\newblock {NeRF-Loc}: Transformer-based object localization within neural radiance fields.
\newblock {\em RAL}, 8(8):5244--5250, 2023.

\bibitem{Takikawa2022VBNF}
Towaki Takikawa, Alex Evans, Jonathan Tremblay, Thomas M{\"{u}}ller, Morgan McGuire, Alec Jacobson, and Sanja Fidler.
\newblock Variable bitrate neural fields.
\newblock In Munkhtsetseg Nandigjav, Niloy~J. Mitra, and Aaron Hertzmann, editors, {\em SIGGRAPH}, pages 41:1--41:9. {ACM}, 2022.

\bibitem{tancik2022block}
Matthew Tancik, Vincent Casser, Xinchen Yan, Sabeek Pradhan, Ben Mildenhall, Pratul~P Srinivasan, Jonathan~T Barron, and Henrik Kretzschmar.
\newblock {Block-NeRF}: Scalable large scene neural view synthesis.
\newblock In {\em CVPR}, pages 8248--8258. IEEE, 6 2022.

\bibitem{Tancik2022blocknerf}
Matthew Tancik, Vincent Casser, Xinchen Yan, Sabeek Pradhan, Ben~P. Mildenhall, Pratul~P. Srinivasan, Jonathan~T. Barron, and Henrik Kretzschmar.
\newblock {Block-NeRF}: Scalable large scene neural view synthesis.
\newblock In {\em CVPR}, pages 8238--8248. {IEEE}, 2022.

\bibitem{tewari2020state}
Ayush Tewari, Ohad Fried, Justus Thies, et~al.
\newblock State of the art on neural rendering.
\newblock In {\em CGF}, volume~39, pages 701--727. Wiley Online Library, Wiley, 5 2020.

\bibitem{tewari2022advances}
Ayush Tewari, Justus Thies, Ben Mildenhall, et~al.
\newblock Advances in neural rendering.
\newblock In {\em CGF}, volume~41, pages 703--735. Wiley Online Library, Wiley, 5 2022.

\bibitem{tretschk2021non}
Edgar Tretschk, Ayush Tewari, Vladislav Golyanik, Michael Zollhöfer, Christoph Lassner, and Christian Theobalt.
\newblock Non-rigid neural radiance fields: Reconstruction and novel view synthesis of a dynamic scene from monocular video.
\newblock In {\em ICCV}, pages 12959--12970. IEEE, 10 2021.

\bibitem{truong2023sparf}
Prune Truong, Marie-Julie Rakotosaona, Fabian Manhardt, and Federico Tombari.
\newblock Sparf: Neural radiance fields from sparse and noisy poses.
\newblock In {\em Proceedings of the IEEE/CVF Conference on Computer Vision and Pattern Recognition}, pages 4190--4200, 2023.

\bibitem{tseng2022cla}
Wei-Cheng Tseng, Hung-Ju Liao, Lin Yen-Chen, and Min Sun.
\newblock {CLA-NeRF}: Category-level articulated neural radiance field.
\newblock {\em arXiv preprint arXiv:2202.00181}, 5 2022.

\bibitem{turki2022mega}
Haithem Turki, Deva Ramanan, and Mahadev Satyanarayanan.
\newblock {Mega-NeRF}: Scalable construction of large-scale {NeRFs} for virtual fly-throughs.
\newblock In {\em CVPR}, pages 12922--12931. IEEE, 6 2022.

\bibitem{verbin2022refnerf}
Dor Verbin, Peter Hedman, Ben Mildenhall, Todd Zickler, Jonathan~T Barron, and Pratul~P Srinivasan.
\newblock {Ref-NeRF}: Structured view-dependent appearance for neural radiance fields.
\newblock In {\em CVPR}, pages 5481--5490. IEEE, 2022.

\bibitem{verbin2024nerf}
Dor Verbin, Pratul~P Srinivasan, Peter Hedman, Ben Mildenhall, Benjamin Attal, Richard Szeliski, and Jonathan~T Barron.
\newblock Nerf-casting: Improved view-dependent appearance with consistent reflections.
\newblock {\em arXiv preprint arXiv:2405.14871}, 2024.

\bibitem{wang2022nerf}
Chen Wang, Xian Wu, Yuan-Chen Guo, Song-Hai Zhang, Yu-Wing Tai, and Shi-Min Hu.
\newblock Nerf-sr: High quality neural radiance fields using supersampling.
\newblock In {\em MM}, pages 6445--6454, 2022.

\bibitem{wang2022r2l}
Huan Wang, Jian Ren, Zeng Huang, Kyle Olszewski, Menglei Chai, Yun Fu, and Sergey Tulyakov.
\newblock {R2l}: Distilling neural radiance field to neural light field for efficient novel view synthesis.
\newblock In {\em ECCV}, pages 612--629. Springer, 2022.

\bibitem{wang2021neus}
Peng Wang, Lingjie Liu, Yuan Liu, Christian Theobalt, Taku Komura, and Wenping Wang.
\newblock Neus: Learning neural implicit surfaces by volume rendering for multi-view reconstruction.
\newblock In {\em NeurIPS}, 2021.

\bibitem{wang2023neus2}
Yiming Wang, Qin Han, Marc Habermann, Kostas Daniilidis, Christian Theobalt, and Lingjie Liu.
\newblock Neus2: Fast learning of neural implicit surfaces for multi-view reconstruction.
\newblock In {\em ICCV}, pages 3295--3306, 2023.

\bibitem{wang2004ssim}
Zhou Wang, Alan~C Bovik, Hamid~R Sheikh, and Eero~P Simoncelli.
\newblock Image quality assessment: from error visibility to structural similarity.
\newblock {\em TIP}, 13(4):600--612, 2004.

\bibitem{wang20224k}
Zhongshu Wang, Lingzhi Li, Zhen Shen, Li Shen, and Liefeng Bo.
\newblock 4k-nerf: High fidelity neural radiance fields at ultra high resolutions.
\newblock {\em arXiv preprint arXiv:2212.04701}, 2022.

\bibitem{wang2021nerf}
Zirui Wang, Shangzhe Wu, Weidi Xie, Min Chen, and Victor~Adrian Prisacariu.
\newblock {NeRF--}: Neural radiance fields without known camera parameters.
\newblock {\em arXiv preprint arXiv:2102.07064}, 2 2021.

\bibitem{wei2021nerfingmvs}
Yi Wei, Shaohui Liu, Yongming Rao, Wang Zhao, Jiwen Lu, and Jie Zhou.
\newblock {NerfingMVS}: Guided optimization of neural radiance fields for indoor multi-view stereo.
\newblock In {\em ICCV}, pages 5610--5619. IEEE, 10 2021.

\bibitem{wu2023nefii}
Haoqian Wu, Zhipeng Hu, Lincheng Li, Yongqiang Zhang, Changjie Fan, and Xin Yu.
\newblock Nefii: Inverse rendering for reflectance decomposition with near-field indirect illumination.
\newblock In {\em CVPR}, pages 4295--4304, 2023.

\bibitem{wu2024neural}
Liwen Wu, Sai Bi, Zexiang Xu, Fujun Luan, Kai Zhang, Iliyan Georgiev, Kalyan Sunkavalli, and Ravi Ramamoorthi.
\newblock Neural directional encoding for efficient and accurate view-dependent appearance modeling.
\newblock {\em arXiv preprint arXiv:2405.14847}, 2024.

\bibitem{Xia-2022-SiNeRF}
Yitong Xia, Hao Tang, Radu Timofte, and Luc~Van Gool.
\newblock {SiNeRF}: Sinusoidal neural radiance fields for joint pose estimation and scene reconstruction.
\newblock In {\em BMVC}, page 131. {BMVA} Press, 2022.

\bibitem{xian2021space}
Wenqi Xian, Jia-Bin Huang, Johannes Kopf, and Changil Kim.
\newblock Space-time neural irradiance fields for free-viewpoint video.
\newblock In {\em CVPR}, pages 9421--9431. IEEE, 6 2021.

\bibitem{Yariv2023baked}
Lior Yariv, Peter Hedman, Christian Reiser, Dor Verbin, Pratul~P. Srinivasan, Richard Szeliski, Jonathan~T. Barron, and Ben Mildenhall.
\newblock {BakedSDF}: Meshing neural sdfs for real-time view synthesis.
\newblock In Erik Brunvand, Alla Sheffer, and Michael Wimmer, editors, {\em SIGGRAPH}, pages 46:1--46:9. {ACM}, 2023.

\bibitem{ye20243d}
Keyang Ye, Qiming Hou, and Kun Zhou.
\newblock 3d gaussian splatting with deferred reflection.
\newblock {\em arXiv preprint arXiv:2404.18454}, 2024.

\bibitem{yen2021inerf}
Lin Yen-Chen, Pete Florence, Jonathan~T Barron, Alberto Rodriguez, Phillip Isola, and Tsung-Yi Lin.
\newblock {iNeRF}: Inverting neural radiance fields for pose estimation.
\newblock In {\em IROS}, pages 1323--1330. IEEE, 9 2021.

\bibitem{yin2023msnerf}
Ze-Xin Yin, Jiaxiong Qiu, Ming-Ming Cheng, and Bo Ren.
\newblock Multi-space neural radiance fields.
\newblock In {\em CVPR}, pages 12407--12416, 2023.

\bibitem{zeng2023mirror}
Junyi Zeng, Chong Bao, Rui Chen, Zilong Dong, Guofeng Zhang, Hujun Bao, and Zhaopeng Cui.
\newblock Mirror-nerf: Learning neural radiance fields for mirrors with whitted-style ray tracing.
\newblock In {\em MM}, 2023.

\bibitem{zhang2021physg}
Kai Zhang, Fujun Luan, Qianqian Wang, Kavita Bala, and Noah Snavely.
\newblock Physg: Inverse rendering with spherical gaussians for physics-based material editing and relighting.
\newblock In {\em CVPR}, pages 5453--5462, 2021.

\bibitem{zhang2020nerf++}
Kai Zhang, Gernot Riegler, Noah Snavely, and Vladlen Koltun.
\newblock Nerf++: Analyzing and improving neural radiance fields.
\newblock {\em arXiv preprint arXiv:2010.07492}, 2020.

\bibitem{zhang2018lpips}
Richard Zhang, Phillip Isola, Alexei~A Efros, Eli Shechtman, and Oliver Wang.
\newblock The unreasonable effectiveness of deep features as a perceptual metric.
\newblock In {\em CVPR}, pages 586--595, 2018.

\bibitem{zhang2022nerfusion}
Xiaoshuai Zhang, Sai Bi, Kalyan Sunkavalli, Hao Su, and Zexiang Xu.
\newblock {NeRFusion}: Fusing radiance fields for large-scale scene reconstruction.
\newblock In {\em CVPR}, pages 5449--5458, 2022.

\bibitem{Zhang2023Nerflets}
Xiaoshuai Zhang, Abhijit Kundu, Thomas Funkhouser, Leonidas Guibas, Hao Su, and Kyle Genova.
\newblock Nerflets: Local radiance fields for efficient structure-aware 3d scene representation from 2d supervision.
\newblock In {\em CVPR}, pages 8274--8284, 2023.

\bibitem{zhang2021nerfactor}
Xiuming Zhang, Pratul~P Srinivasan, Boyang Deng, Paul Debevec, William~T. Freeman, and Jonathan~T. Barron.
\newblock Nerfactor: Neural factorization of shape and reflectance under an unknown illumination.
\newblock {\em TOG}, 40(6):1--18, 2021.

\bibitem{zhang2022modeling}
Yuanqing Zhang, Jiaming Sun, Xingyi He, Huan Fu, Rongfei Jia, and Xiaowei Zhou.
\newblock Modeling indirect illumination for inverse rendering.
\newblock In {\em CVPR}, pages 18643--18652, 2022.

\bibitem{zhang2023nemf}
Youjia Zhang, Teng Xu, Junqing Yu, Yuteng Ye, Yanqing Jing, Junle Wang, Jingyi Yu, and Wei Yang.
\newblock Nemf: Inverse volume rendering with neural microflake field.
\newblock In {\em ICCV}, pages 22919--22929, 2023.

\bibitem{zhao2022factorized}
Boming Zhao, Bangbang Yang, Zhenyang Li, Zuoyue Li, Guofeng Zhang, Jiashu Zhao, Dawei Yin, Zhaopeng Cui, and Hujun Bao.
\newblock Factorized and controllable neural re-rendering of outdoor scene for photo extrapolation.
\newblock In {\em MM}, pages 1455--1464, 2022.

\bibitem{zhu2023nicer}
Zihan Zhu, Songyou Peng, Viktor Larsson, Zhaopeng Cui, Martin~R Oswald, Andreas Geiger, and Marc Pollefeys.
\newblock {NICER-SLAM}: Neural implicit scene encoding for {RGB SLAM}.
\newblock {\em arXiv preprint arXiv:2302.03594}, 2023.

\bibitem{zhu2022nice}
Zihan Zhu, Songyou Peng, Viktor Larsson, Weiwei Xu, Hujun Bao, Zhaopeng Cui, Martin~R Oswald, and Marc Pollefeys.
\newblock {Nice-SLAM}: Neural implicit scalable encoding for {SLAM}.
\newblock In {\em CVPR}, pages 12786--12796. IEEE, 6 2022.

\end{thebibliography}
}

\end{document}